\documentclass{article}

\PassOptionsToPackage{numbers, compress}{natbib}
\usepackage[preprint]{neurips_2024}

\usepackage[utf8]{inputenc} % allow utf-8 input
\usepackage[T1]{fontenc}    % use 8-bit T1 fonts
\usepackage{hyperref}       % hyperlinks
\usepackage{url}            % simple URL typesetting
\usepackage{booktabs}       % professional-quality tables
\usepackage{amsfonts}       % blackboard math symbols
\usepackage{nicefrac}       % compact symbols for 1/2, etc.
\usepackage{microtype}      % microtypography
\usepackage{xcolor}         % colors
\definecolor{LightGray}{gray}{0.9}

\usepackage{wrapfig}
\usepackage{amsmath, bm, amssymb, amsthm}
\usepackage{graphicx}
\usepackage{algorithm} 
\usepackage{subcaption}
\usepackage{caption}
\usepackage{algcompatible}

\usepackage{listings}

\definecolor{codegreen}{rgb}{0,0.6,0}
\definecolor{codegray}{rgb}{0.5,0.5,0.5}
\definecolor{codepurple}{rgb}{0.58,0,0.82}
\definecolor{backcolour}{rgb}{0.95,0.95,0.92}

\lstdefinestyle{mystyle}{
    backgroundcolor=\color{backcolour},   
    commentstyle=\color{codegreen},
    keywordstyle=\color{magenta},
    numberstyle=\tiny\color{codegray},
    stringstyle=\color{codepurple},
    basicstyle=\ttfamily\footnotesize,
    breakatwhitespace=false,         
    breaklines=true,                 
    captionpos=b,                    
    keepspaces=true,                 
    numbers=left,                    
    numbersep=5pt,                  
    showspaces=false,                
    showstringspaces=false,
    showtabs=false,                  
    tabsize=2
}

\title{Equivariant Blurring Diffusion \\ for Hierarchical Molecular Conformer Generation}

\author{%
  Jiwoong Park, Yang Shen \\
  Department of Electrical and Computer Engineering\\
  Texas A\&M University\\
  \texttt{ptywoong@gmail.com, yshen@tamu.edu} \\
}

\begin{document}

\maketitle

\begin{abstract}
How can diffusion models process 3D geometries in a coarse-to-fine manner, akin to our multiscale view of the world?
In this paper, we address the question by focusing on a fundamental biochemical problem of generating 3D molecular conformers conditioned on molecular graphs in a multiscale manner. 
Our approach consists of two hierarchical stages: i) generation of coarse-grained fragment-level 3D structure from the molecular graph, and ii) generation of fine atomic details from the coarse-grained approximated structure while allowing the latter to be adjusted simultaneously.
For the challenging second stage, which demands preserving coarse-grained information while ensuring SE(3) equivariance, we introduce a novel generative model termed \textit{Equivariant Blurring Diffusion} (EBD), which defines a forward process that moves towards the fragment-level coarse-grained structure by blurring the fine atomic details of conformers, and a reverse process that performs the opposite operation using equivariant networks.
We demonstrate the effectiveness of EBD by geometric and chemical comparison to state-of-the-art denoising diffusion models on a benchmark of drug-like molecules.
Ablation studies draw insights on the design of EBD by thoroughly analyzing its architecture, which includes the design of the loss function and the data corruption process. Codes are released at \url{https://github.com/Shen-Lab/EBD}.

\end{abstract}
\section{Introduction}
\begin{wrapfigure}[16]{r}{0.48\textwidth}
\vspace{-20pt}
  \centering
  \includegraphics[trim={0 19cm 0 0}, clip, width=0.48\textwidth]{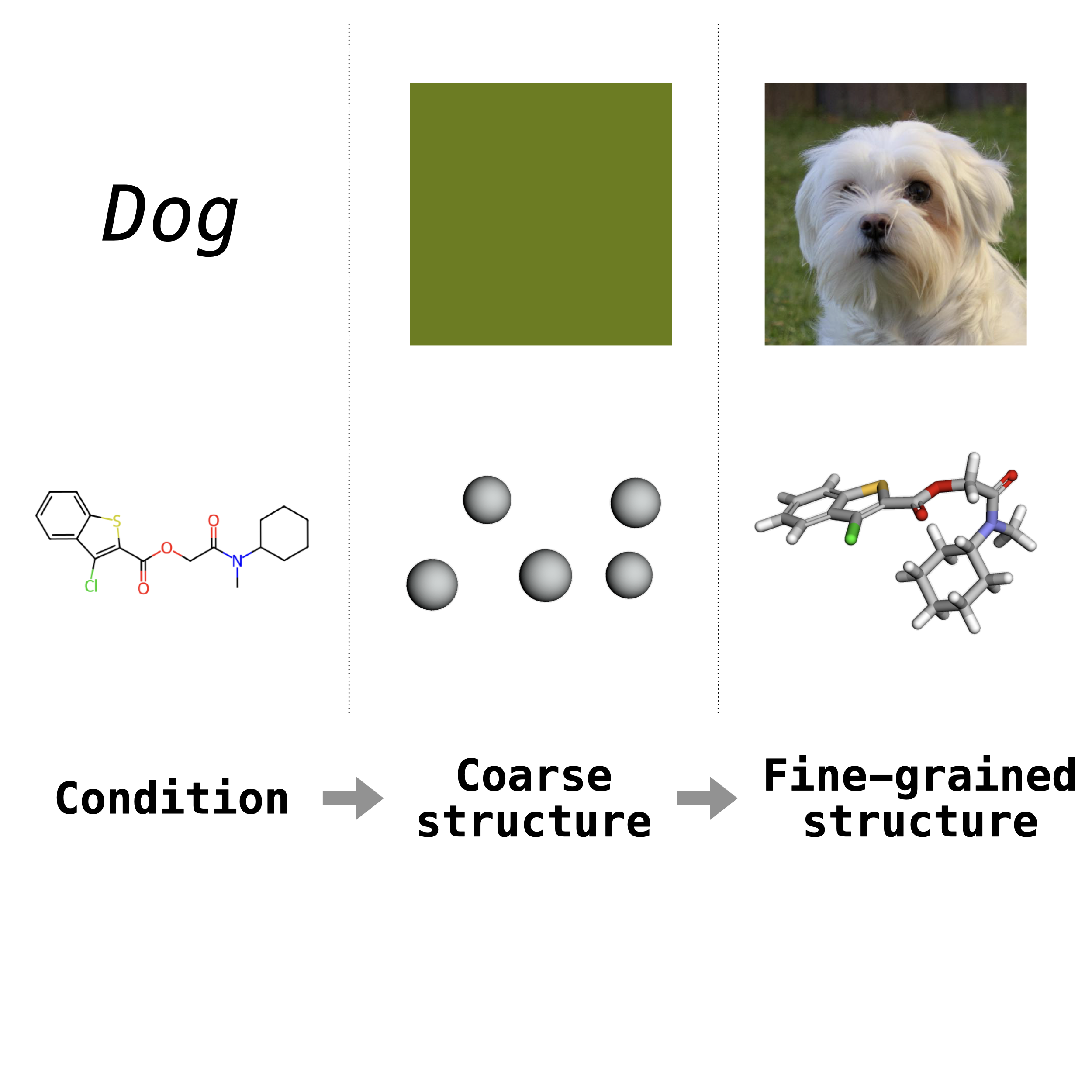}
  \caption{Blurring diffusion generative processes on image \cite{rissanen2022generative} and molecular conformer.
  % \yy{Suggestion: This figure can be put in the first page right below abstract as a teaser, which directly explains the core idea for reviewer at ther first sight.}
  }
  \label{fig:intro}
\end{wrapfigure} 
% \begin{center}
% \textit{How can diffusion models generalize to the coarse-to-fine framework for 3D geometry?}
% \end{center}

The advancement of generative models to understand the multiscale properties of objects facilitates their application across a range of granularity levels, transcending individual scales. To enable generative models in processing multiscale structures, there has been a surge in hierarchical design methodologies across multiple domains, spanning from images \cite{menick2018generating, razavi2019generating} to speech \cite{hsu2018hierarchical, hono2020hierarchical}. These methods initially capture coarse-grained structures and subsequently generate finer details.

In the field of computer vision, recent efforts \cite{ho2022cascaded, rissanen2022generative, bansal2023cold, Giannis2023soft} have yielded successful designs of coarse-to-fine generative frameworks for 2D pixels of images, leveraging denoising diffusion models that corrupt and restore the data by adding and removing noises \cite{sohl2015deep, song2020denoising, song2020score, ho2020denoising, kingma2021variational}. Notably, \cite{rissanen2022generative} generates images from blurred prior distributions (average of pixel intensities) motivated from the heat equation of partial differential equations as Fig. \ref{fig:intro}.

In the field of biochemistry and drug discovery, however,  denoising diffusion models for 3D conformers of stable molecular structures have not yet taken advantage of coarse-to-fine  multiscale frameworks. Current methods either disregard the scale hierarchy \cite{shi2021learning, xu2022geodiff, hoogeboom2022equivariant, jing2022torsional, xu2023geometric} or consider that in very limited ways \cite{qiang2023coarse,reidenbach2023coarsenconf}.  For instance, within the recent hierarchical method of unconditional conformer generation \cite{qiang2023coarse}, a denoising diffusion model \cite{hoogeboom2022equivariant} is solely applied to generation of coarse-grained structure, without extending to coarse-to-fine generation.

The primary bottleneck in extending denoising diffusion models for molecular conformers  to hierarchical designs is that random noise corrupts not only fine atomic details but also structural information of coarse-grained structures indiscriminately. To tackle this challenging problem, we exploit \textit{fragments} that are frequently occurring substructures or functional groups in 2D molecular graphs. These fragments can be a promising candidate for the coarse-grained structural information in 3D geometry. Introducing fragments divides the generation process into two stages: i) generating coarse-grained structures represented by fragments, and ii) restoring fine atomic details from fragment structures. In the first stage of generating fragment coordinates from molecular graphs, we efficiently generate approximations of fragment structures comprising the center of mass and attributes of each fragment from a cheminformatics tool.

For the challenging second step of coarse-to-fine generation, we propose a novel diffusion model, \textit{Equivariant Blurring Diffusion} (EBD) detailed as follows. Motivated from the blurring corruption of the heat equation \cite{rissanen2022generative}, we design EBD to generate 3D molecular conformers from coarse-grained fragment approximated  structures as Fig. \ref{fig:intro}, rather than from random noise. In our design of EBD, the forward process moves atom positions of conformers towards the center of mass of their respective fragments, while the reverse process restores full-atom details from the prior distribution of the 3D fragment structure. The blurring schedule we designed for EBD allows the diffusion model to focus on restoring fine atomic details while retaining coarse-grained information throughout the entire generative process. We validated our coarse-to-fine EBD model using a benchmark of drug-like molecules. We obtained superior results in conformer generation compared to the denoising diffusion model, even with 100 times fewer diffusion time steps. 

% tailored to our coarse-to-fine generation problem 
% to hierarchical generation of molecular conformer using fragments. We first efficiently generate approximations of fragment structures that consist of the center of mass and attributes of each fragment from a cheminformatics tool. Then, a diffusion model generates full-atom details from fragment structures. 
% Thus, we designed 
% Motivated by these works, we introduces  that generates molecular conformers from fragment structures as Fig. \ref{fig:intro}.  

The major contribution of this paper can be summarized as follows:
\begin{itemize}
    \item 
    % We propose a hierarchical framework for molecular conformer generation. 
     We design EBD which generates atomic details from coarse-grained estimation of fragment structures using equivariant networks, motivated by the blurring corruption of heat equation.
    \item We propose a novel blurring scheduler and a revised loss function that significantly impacts  performance, instead of directly applying those of existing image blurring diffusion model.
    \item We conduct a thorough analysis of the effects of fragment granularity, data corruption methods, and loss reformulation. We obtained geometrically and chemically more plausible conformers compared to state-of-the-art denoising diffusion models.
    % Even with the denoising diffusion model starts from fragment structure-encoded prior distribution, our model maintain s superiority.
\end{itemize}

\section{Background}
\subsection{Blurring diffusion}
The denoising diffusion models \cite{sohl2015deep, song2020denoising, song2020score, ho2020denoising, kingma2021variational}, which corrupt data by adding random noise and generate data through denoising, have significantly advanced across diverse domains \cite{vahdat2022lion, vahdat2021score, saharia2022photorealistic}. Recently, a few works \cite{bansal2023cold, rissanen2022generative, Giannis2023soft, hoogeboom2022blurring} have introduced data corruption into the design space of diffusion models \cite{karras2022elucidating}, going beyond random noise corruption in the vision domain. 

Inverse Heat Dissipation Model (IHDM) \cite{rissanen2022generative} proposed a coarse-to-fine generation in the pixel space. Their forward process follows a partial differential equation of heat dissipation on grids: 
\begin{equation}
    \frac{\partial}{\partial t}\mathbf{x}(i, j, t) = \Delta \mathbf{x}(i, j, t),
\end{equation}
where $\mathbf{x}$ represents the data on the grid and $\Delta$ is the Laplacian operator. IHDM derived the solution of this equation at time step $t$, $\mathbf{x}_t$, using eigendecomposition of $\Delta$ as:
\begin{equation} \label{eq:ihdm_fwd}
    \mathbf{x}_t = \mathbf{B}_t\mathbf{x}_0 = \mathbf{V} \exp (- \mathbf{\Lambda}t) \mathbf{V}^T \mathbf{x}_0,
\end{equation}
where $\mathbf{V}^T$ and $\mathbf{\Lambda}$ are discrete cosine transform and a diagonal matrix whose elements are eigenvalues of $\Delta$, respectively. As $t \rightarrow T$, the eigenbasis of eigenvalue $0$ only remains and this leads to the convergence of pixel intensities to their average value. Based upon this blurring process, IHDM defined a forward process as:
\begin{equation} 
    q(\mathbf{x}_t \vert \mathbf{x}_0) = \mathcal{N} (\mathbf{x}_t \vert  \mathbf{B}_t \mathbf{x}_0, \sigma^2 \mathbf{I}),
\end{equation}
which means that the state at $t$ is equal to the data blurred until $t$ with small amount of noise. Note that the function of data corruption $\mathbf{B}_t$ was defined at a spectral space of eigenvalues $\mathbf{\Lambda}$. Then, the reverse generative process was defined to deblur each state:
\begin{equation}
    p_{\theta}(\mathbf{x}_{t-1} \vert \mathbf{x}_t) = \mathcal{N} (\mathbf{x}_{t-1} \vert \mu_{\theta}(\mathbf{x}_t, t), \delta^2 \mathbf{I}),
\end{equation}
where the mean at $t-1$ is the result of a deblurring network $\mu_{\theta}$ and $\delta$ is the small amount of standard deviation for noise. As $t$ approaches $0$, $\mu_\theta$ gradually restores fine details from coarse-grained information about pixel intensities by effectively deblurring state values.
% incrementally deblurs the state value from coarse-grained information about pixel intensities and gradually restores fine details. 
The loss was defined to minimize the distance between the result of deblurring network and less blurred state at randomly sampled $t$ as:
\begin{equation} \label{eq:ihdm_loss}
    L_{t-1} = \mathbb{E}_{t, \mathbf{x}_0, \mathbf{x}_t} [\| \mathbf{B}_{t-1} \mathbf{x}_0 - \mu_{\theta}(\mathbf{x}_t, t) \|^2].
\end{equation}
IHDM was evaluated on image generation task using FID score \cite{heusel2017gans}, but its performance lagged behind that of denoising diffusion models. For instance, IHDM achieves an FID score of $18.96$ while DDPM \cite{ho2020denoising} have $3.17$ on CIFAR-10 \cite{krizhevsky2009learning}.

% \begin{align}
%     q(u_{1:K} \vert u_0) &= \Pi_{k=1}^K q(u_k \vert u_{0}) \nonumber \\ 
%     &= \Pi_{k=1}^K \mathcal{N}(u_k \vert \mathbf{F}(t_k)u_0, \sigma^2I) \\
%     &= \Pi_{k=1}^K \mathcal{N}(u_k \vert \mathbf{V}\exp(-\mathbf{\Lambda} t_k)\mathbf{V}^Tu_0, \sigma^2 I), \nonumber
% \end{align}

% \begin{align}
%     p_{\theta}(u_{0:K}) &= p(u_K) \Pi_{k=1}^K p_{\theta}(u_{k-1} \vert u_k) \nonumber \\ 
%     &= p(u_K) \Pi_{k=1}^K \mathcal{N}(u_{k-1} \vert \mu_{\theta}(u_k, k), \delta^2 I) \\ 
%     &= p(u_K) \Pi_{k=1}^K \mathcal{N}(u_{k-1} \vert u_k + f_{\theta}(u_k, k), \delta^2 I). \nonumber
% \end{align}

\subsection{Equivariance}
% Our geometry generation is not in the pixel space, which poses unique needs of SE(3)-equivariance.  
In this work, we consider the SE(3) group to address the roto-translational equivariance of molecular conformers \cite{kohler2020equivariant, hoogeboom2022equivariant, xu2022geodiff}. A function $f$ is equivariant to a group $\mathcal{G}$ if $T_g(f(\mathbf{x})) = f(S_g(\mathbf{x}))$ holds for all $g \in \mathcal{G}$, where $T_g, S_g$ are transformations of the group element $g$.  In our coarse-to-fine generative framework, the invariant prior distribution of coarse-grained structure represents the coordinates of fragments. Therefore, the design of the transition distribution and the loss function in our diffusion model need to ensure that the generated likelihood is invariant, so that the generated conformers are not affected by rotation or translation.

\section{Methods} \label{sec:methods}
\subsection{Problem definition}
Suppose that we have a molecular graph $G$ whose nodes $\mathcal{V}$ are $n$ atoms with  SE(3)-invariant features $\mathbf{h}^\text{a} \in \mathbb{R}^{n \times d}$ and edges $\mathcal{E}$ are inter-atomic bonds. Our objective is to generate an ensemble of 3D molecular conformers $\mathbf{x}^\text{a} \in \mathbb{R}^{n \times 3}$ given $G$. Our hierarchical approach is in two stages. i) $p(\mathbf{x}^\text{f} \vert G)$: generating a coarse-grained 3D structure of fragment coordinates $\mathbf{x}^\text{f} \in \mathbb{R}^{m \times 3}$ from $G$ which was decomposed into $m$ fragments, and ii) $p(\mathbf{x}^\text{a} \vert \mathbf{x}^\text{f}, G)$: the diffusion model generating fine atomic details $\mathbf{x}^\text{a} \in \mathbb{R}^{n \times 3}$ 
% and simultaneously correcting coarse-grained fragment structures, 
conditioned on the generated fragment structure $\mathbf{x}^\text{f}$. 
% Superscripts ``a'' and ``f'' indicate the fine-grained atom-level and the coarse-grained fragment-level, respectively. 
To map  between atoms and their respective fragments, we defined a mapping matrix $\mathbf{M} \in \mathbb{R}^{n \times m}$ with $\mathbf{M}_{\mathtt{ik}} = 1$ if the $i$-th atom belongs to the $k$-th fragment and $0$ otherwise. $\mathbf{M} \mathbf{x}^\text{f}$ makes each atom located at its respective fragment. On the other hand, $\mathbf{M}^{\dagger} \mathbf{x}^\text{a}$ results in fragment coordinates being the average of the coordinates of its constituent atoms, where $\mathbf{M}^{\dagger}$ is a pseudoinverse matrix of $\mathbf{M}$ ($\mathbf{M}^\dagger \mathbf{M} = \mathbf{I}$).

\subsection{Fragmentation and 3D fragment structures}
We decompose a molecule $G = (\mathcal{V}, \mathcal{E})$ into $m$  non-overlapping fragments $\{S_k\}_{k=1}^m$, where $S_k = (\mathcal{V}_k, \mathcal{E}_k)$ and $\mathcal{V} = \bigcup_{k=1}^m \mathcal{V}_k, \mathcal{E} = \bigcup_{k=1}^m \mathcal{E}_k$ using Principal Subgraph (PS) \cite{kong2022molecule}. Starting from all unique atoms in the fragment vocabulary $\mathcal{S}$, PS iteratively merges neighboring fragments. The most frequent fragment among the newly merged fragments was added to the vocabulary at each iteration, which was repeated until the desired size of the vocabulary was reached. The smaller the size of fragment vocabulary, the finer fragments and detailed coarse-grained structures can be obtained. After fragmentation was finished, from the relation between $\{\mathcal{V}_k\}_{k=1}^m$ and $\mathcal{V}$, the mapping matrix $\mathbf{M}$ can be constructed.

To generate the initial coordinates of fragments, we utilize RDKit distance geometry \cite{crippen1988distance, landrum2013rdkit}, an efficient cheminformatics tool, instead of training an additional deep generative model. After generating initial atom coordinates $\hat{\mathbf{x}}^\text{a} \sim p_{\text{RDKit}}(\mathbf{x}^\text{a})$, we define the initial fragment coordinates $\mathbf{x}^\text{f}$ as averages of their constituent atom coordinates, $\mathbf{M}^{\dagger} \hat{\mathbf{x}}^\text{a}$. 
Since the atom coordinates generated by RDKit are approximations of the ground truth conformer, the resulting fragment coordinates are also an approximation (thus need to be adjusted in the next stage in Sec.~\ref{sec:methods-ebd}), which we denote as $\hat{\mathbf{x}}^\text{f} \sim p_{\text{RDKit}}(\mathbf{x}^\text{f})$. For fragment features $\mathbf{h}^\text{f} \in \mathbb{R}^{m \times 3}$, we define a $3$-dimensional vector as a frequency histogram of its constituent atom types based on their chemical properties, including hydrophobicity, hydrogen bond center, and negative charge center following \cite{qiang2023coarse}.

The processes of fragment structure generation are illustrated in the step 1 of Fig. \ref{fig:model}. Note that every step in this subsection does not harm the efficiency of our framework, as they can be completed before training our diffusion model and the process itself is efficient. To generate $5$ distinct fragment coordinates $\hat{\mathbf{x}}^\text{f}$ for each of the $45,000$ molecules in the training and validation set of the GEOM-Drug benchmark \cite{axelrod2022geom}, it took $38$ hours, averaging $3.04$ seconds per molecule. The details of fragmentation and fragment vocabulary are demonstrated in Appendix \ref{apdx:subsec_data}.

\subsection{Equivariant blurring diffusion}\label{sec:methods-ebd}
\begin{figure}
  \centering
  \includegraphics[trim={1cm 18cm 10cm 0}, clip, width=\textwidth]{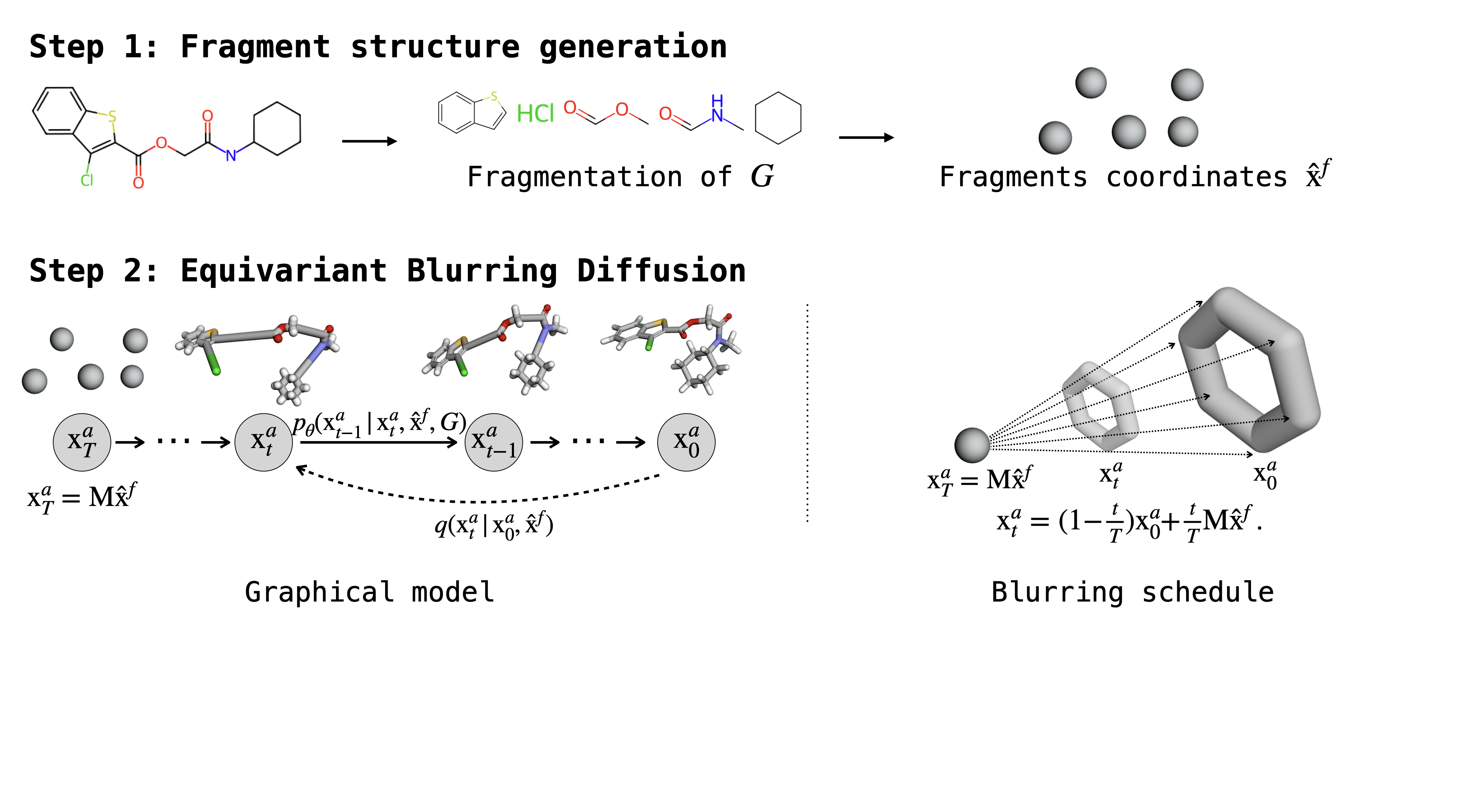}
  \vspace{-15pt}
  \caption{Our hierarchical molecular conformer generation framework. We first decompose a molecular graph $G$ into fragments and generate fragment coordinates $\hat{\mathbf{x}}^f$. Then, conditioned on  $\hat{\mathbf{x}}^f$ and $G$, Equivariant Blurring Diffusion generates atom-level fine details using the blurring schedule.
  % was defined in the spatial domain. 
  % (De)blurring can be either spatial domain (as visualized and adopted in the Euclidean space) or spectral (tested but not adopted). 
  }
  \label{fig:model}
  \vspace{-12pt}
\end{figure}
In this subsection, we elaborate on the design of our diffusion model, \textit{Equivariant Blurring Diffusion} (EBD), drawing inspiration from the principles of the heat equation. This model is designed to generate fine details of conformers $\mathbf{x}^\text{a}$, starting from a coarse-grained, approximate structure $\hat{\mathbf{x}}^\text{f}$ and a molecular graph $G$.  We introduce a forward process and a data corruption function of blurring process in Sec. \ref{subsec:fwd}, a reverse process and a deblurring network to reach an SE(3)-invariant likelihood in Sec. \ref{subsec:rev}, and a definition and reparameterization of an SE(3)-invariant loss function in Sec. \ref{subsec:loss}. The overall scheme of EBD is illustrated in the step 2 of Fig. \ref{fig:model}.

\subsubsection{Forward process and blurring schedule} \label{subsec:fwd}
We define the data corruption of the forward process as a blurring operation that gradually shifts ground truth atom positions $\mathbf{x}_0^\text{a} \sim q(\mathbf{x}_0^\text{a})$ to their corresponding fragment coordinates:
\begin{equation} 
    q(\mathbf{x}^\text{a}_t \vert \mathbf{x}^\text{a}_0, \hat{\mathbf{x}}^\text{f}) = \mathcal{N} (\mathbf{x}^\text{a}_t \vert  f_{\mathbf{B}}(\mathbf{x}^\text{a}_0, \hat{\mathbf{x}}^\text{f}, t),  \sigma^2 \mathbf{I}),
    \label{eq:for}
\end{equation}
where $f_{\mathbf{B}}$ is a deterministic blurring operator. Consequently, every atom will be positioned according to its fragment coordinates $\mathbf{M}\hat{\mathbf{x}}^\text{f}$ in the prior fragment structure distribution.

When defining $f_{\mathbf{B}}$ for the forward process, we cannot directly adopt the spectral blurring operator $\mathbf{B}_t = \mathbf{V} \exp (- \mathbf{\Lambda}t) \mathbf{V}^T $ of IHDM \cite{rissanen2022generative} in Eq. (\ref{eq:ihdm_fwd}) for the two reasons: i) For a single molecule, we need to calculate and decompose the fragment graph Laplacian $\{\mathbf{V}_k \mathbf{\Lambda}_k \mathbf{V}_k^T \}_{k=1}^m$ for each fragment $S_k = (\mathcal{V}_k, \mathcal{E}_k)$, unlike a single Laplacian operator per image in IHDM. Given the varying sizes and structures across fragments, it becomes challenging to uniformly adjust the movement of atoms across all fragments using a function of $\{\mathbf{\Lambda}_k\}_{k=1}^m$ in spectral space. ii) As $t \rightarrow T$, the ground truth atom coordinates $\mathbf{x}_0^\text{a}$ will converge to the ground truth scaffold structure $\mathbf{x}^\text{f} = \mathbf{B}_T \mathbf{x}_0^\text{a}$ by spectral graph theory \cite{chung1997spectral}. However, there exists a mismatch between the ground truth coordinates $\mathbf{x}^\text{f}$ and the approximation coordinates $\hat{\mathbf{x}}^\text{f}$ from RDKit in the generative processes. This distributional shift of the fragment structure can potentially harm the performance during the inference.

To circumvent these issues, we transition the space of the blurring operator from spectral domain to spatial domain while retaining the essence of the blurring process. We define $f_{\mathbf{B}}$ as a linear interpolation between $\mathbf{M}\hat{\mathbf{x}}^\text{f}$ and $\mathbf{x}^\text{a}_0$ in Euclidean space: 
\begin{equation} \label{eq:blur_sch}
    f_{\mathbf{B}}(\mathbf{x}^\text{a}_0, \hat{\mathbf{x}}^\text{f}, t) = (1 - \frac{t}{T}) \mathbf{x}^\text{a}_0 + \frac{t}{T} \mathbf{M}\hat{\mathbf{x}}^\text{f}.
\end{equation}
As $t$ progresses from $0$ to $T$, the atom coordinates $\mathbf{x}^\text{a}_t$ will gradually converge to the fragment structure $\mathbf{M}\hat{\mathbf{x}}^\text{f}$, allowing for uniform adjustment of atom movement. Additionally, we can mitigate the need for excessive eigendecomposition of the fragment graph Laplacian. The example of our blurring schedule on a single fragment is depicted in the step 2 of  Fig. \ref{fig:model}.

\subsubsection{Reverse process and deblurring networks} \label{subsec:rev}
The aim of the reverse process is to generate fine details at the atom-level from a prior distribution of 3D fragment structure $p(\mathbf{x}^\text{a}_T) = \mathcal{N}(\mathbf{x}^\text{a}_T \vert \mathbf{M}\hat{\mathbf{x}}^\text{f}, \delta^2 \mathbf{I})$ that is roto-translational invariant to the group. Drawing upon proofs regarding the conditions for an invariant likelihood \cite{kohler2020equivariant, xu2022geodiff}, we develop the deblurring process on the zero center-of-mass subspace using equivariant transition distributions:
\begin{equation}
    p_{\theta}(\mathbf{x}^\text{a}_{t-1} \vert \mathbf{x}^\text{a}_t, \hat{\mathbf{x}}^\text{f}, G) = \mathcal{N} (\mathbf{x}^\text{a}_{t-1} \vert \mu_{\theta}(\mathbf{x}^\text{a}_t, \hat{\mathbf{x}}^\text{f}, G, t), \delta^2 \mathbf{I}),
    \label{eq:rev}
\end{equation}
where $\mu_{\theta}$ is a parameterized mean function consisting of a deblurring network. To ensure equivariance in the transition distribution, we devise $\mu_{\theta}$ inspired by equivariant networks \cite{satorras2021n}. Our equivariant deblurring network updates invariant features of fragments and atoms $ \mathbf{h}^\text{f}, \mathbf{h}^\text{a},$ (Eqs. (\ref{eq:inv_frag}, \ref{eq:inv_atom})), and the equivariant coordinates of atoms $\mathbf{x}^\text{a}$ (Eq. (\ref{eq:eq_atom})) by leveraging the hierarchical relationship between atoms and fragments. Let the $i$-th atom $\mathbf{x}_{\mathtt{i}}^\text{a}$ belongs to the $k$-th fragment $\mathbf{x}_{\mathtt{k}}^\text{f}$, then the $l$-th layer of equivariant deblurring networks for fragment- and atom-level message passing and feature updates is defined as follows:
\begin{align} 
    \mathbf{m}^\text{f}_{\mathtt{ij}} &= \phi_m^\text{f} (\mathbf{h}_{\mathtt{i}}^{\text{f},l}, \mathbf{h}_{\mathtt{j}}^{\text{f},l}, \|\mathbf{x}^\text{f}_{\mathtt{i}} - \mathbf{x}^\text{f}_{\mathtt{j}}\|), & \mathbf{h}_{\mathtt{i}}^{\text{f},l+1} &= \phi_h^\text{f} (\mathbf{h}_{\mathtt{i}}^{\text{f},l}, \sum\nolimits_{\mathtt{{j}} \in N(\mathbf{x}_{\mathtt{i}}^\text{f})} \mathbf{m}_{\mathtt{ij}}^\text{f}, \mathbf{h}^{\text{a},l}), \label{eq:inv_frag}\\
    \mathbf{m}^\text{a}_{\mathtt{ij}} &= \phi_m^\text{a} (\mathbf{h}_{\mathtt{i}}^{\text{a},l}, \mathbf{h}_{\mathtt{j}}^{\text{a},l}, \|\mathbf{x}_{\mathtt{i}}^{\text{a},l} - \mathbf{x}_{\mathtt{j}}^{\text{a},l}\|, e^\text{a}_{\mathtt{ij}}), & \mathbf{h}_{\mathtt{i}}^{\text{a},l+1} &= \phi_h^\text{a} (\mathbf{h}_{\mathtt{i}}^{\text{a},l}, \sum\nolimits_{\mathtt{j} \in N(\mathbf{x}_{\mathtt{i}}^\text{a})} \mathbf{m}_{\mathtt{ij}}^\text{a}, \mathbf{h}^{\text{f},l+1}),\label{eq:inv_atom}
\end{align} 
\vspace{-20pt}
\begin{align}
\begin{split}
\mathbf{x}_{\mathtt{i}}^{\text{a},l+1} = \mathbf{x}_{\mathtt{i}}^{\text{a},l} &+ \sum_{\mathtt{j} \in N(\mathbf{x}_{\mathtt{i}}^\text{a})} \frac{\mathbf{x}_{\mathtt{i}}^{\text{a},l} - \mathbf{x}_{\mathtt{j}}^{\text{a},l}}{d_{\mathtt{ij}}^{\text{a},l} + 1} \phi_x^\text{a}( \mathbf{h}_{\mathtt{i}}^{\text{a},l+1}, \mathbf{h}_{\mathtt{j}}^{\text{a},l+1}, \mathbf{m}^\text{a}_{\mathtt{ij}},  e^\text{a}_{\mathtt{ij}}) \\ 
&+ \frac{\mathbf{x}_{\mathtt{i}}^{\text{a},l} - \mathbf{x}_{\mathtt{k}}^\text{f}}{\|\mathbf{x}_{\mathtt{i}}^{\text{a},l} - \mathbf{x}_{\mathtt{k}}^\text{f}\| + 1} \phi_x^\text{f}(\mathbf{h}_{\mathtt{i}}^{\text{a},l+1},  \mathbf{h}_{\mathtt{k}}^{\text{f},l+1}, \|\mathbf{x}_{\mathtt{i}}^{\text{a},l} - \mathbf{x}_{\mathtt{k}}^\text{f}\|), \label{eq:eq_atom}
\end{split}
\end{align}
where $\mathbf{x}_{\mathtt{k}}^\text{f}$ is the $k$-th row of $\mathbf{M}^\dagger \mathbf{x}^\text{a}_t$, $\phi$ are multilayer perceptrons, $e_{\mathtt{ij}}^\text{a}$ are inter-atomic bond types, and $d_{\mathtt{ij}}^{\text{a}, l} = \|\mathbf{x}_{\mathtt{i}}^{\text{a}, l} - \mathbf{x}_{\mathtt{j}}^{\text{a}, l}\|$ are inter-atomic distances. We consider a complete graph for fragment-level interactions and expand the edge set by incorporating multi-hop and radius neighbors for atom-level interactions. The details of the deblurring networks are provided in the Appendix \ref{apdx:networks}. 
% where subscripts with brackets indicate components of the corresponding vector (for instance, $\mathbf{x}^\text{f}_{[k]}$ is the $k$-th row of $\mathbf{M}^\dagger \mathbf{x}^\text{a}_t$), superscripts $l$ indicate layer index, 

\subsubsection{Training} \label{subsec:loss}
Following Eq. (\ref{eq:ihdm_loss}) of IHDM \cite{rissanen2022generative}, our loss of previous deblurred state estimation can be defined as:
\begin{equation} \label{eq:loss_prev}
    L_{t-1} = \mathbb{E}_{t, \mathbf{x}^\text{a}_0, \mathbf{x}^\text{a}_t, \hat{\mathbf{x}}^\text{f} } [\|  f_{\mathbf{B}}(\mathbf{x}^\text{a}_0, \hat{\mathbf{x}}^\text{f}, t-1) - \rho \big(\mu_{\theta}(\mathbf{x}^\text{a}_t, \hat{\mathbf{x}}^\text{f}, G, t)\big) \|^2],
\end{equation}
where $\rho$ is the Kabsch algorithm \cite{kabsch1976solution} to obtain the optimal rotation matrix for alignment. Through alignment $\rho$ between the prediction from $\mu_\theta$ and less blurred state $f_{\mathbf{B}}(\mathbf{x}^\text{a}_0, \hat{\mathbf{x}}^\text{f}, t-1)$ after translating both terms to the zero center-of-mass subspace, the loss function becomes invariant to the SE(3)-transformation of the prediction. 
% We applied  to $\rho$ to obtain the optimal rotation matrix after translating both terms to the zero center-of-mass subspace.

However, we empirically observed that this previous state estimator generates unsatisfactory conformers, similar to the unsatisfactory FID scores observed in image generation of IHDM \cite{rissanen2022generative}. We conjectured the reason as the model limited to learn the locally small steps towards the ground truth distribution at each time step \cite{Giannis2023soft}. Thus, we reparameterize $\mu_{\theta}(\mathbf{x}^\text{a}_t, \hat{\mathbf{x}}^\text{f}, G, t)$ as $(1 - \frac{t-1}{T}) f_{\theta}(\mathbf{x}^\text{a}_t, G, t) + \frac{t-1}{T} \mathbf{M}\hat{\mathbf{x}}^\text{f}$ to make the deblurring network estimates the ground truth state $\mathbf{x}^\text{a}_0$ instead of the previous less blurred state via neural networks $f_{\theta}$:
\begin{align}
    L_{t-1} &= \mathbb{E}_{t, \mathbf{x}^\text{a}_0, \mathbf{x}^\text{a}_t, \hat{\mathbf{x}}^\text{f}} [\|  f_{\mathbf{B}}(\mathbf{x}^\text{a}_0, \hat{\mathbf{x}}^\text{f}, t-1) - \rho \big((1 - \dfrac{t-1}{T}) f_{\theta}(\mathbf{x}^\text{a}_t, G, t) + \dfrac{t-1}{T} \mathbf{M}\hat{\mathbf{x}}^\text{f} \big) \|^2] \\
    & \approx \mathbb{E}_{t, \mathbf{x}^\text{a}_0, \mathbf{x}^\text{a}_t, \hat{\mathbf{x}}^\text{f}} [\| \mathbf{x}^\text{a}_0 - \rho \big( f_{\theta}(\mathbf{x}^\text{a}_t, G, t) \big) \|^2].   \label{eq:loss_gt}
\end{align}
The derivation of the new loss 
% including the approximation due to simplification (discarding time-dependent weights), 
is detailed in Appendix \ref{apdx:loss}. By loss reparameterization, $\rho$ aligns the prediction to the ground truth state. At time step $t$ of the sampling process, after estimating ground truth $\tilde{\mathbf{x}}^\text{a}_0$ from $\mathbf{x}^\text{a}_t$, the next state $\mathbf{x}^\text{a}_{t-1}$ is computed from a deterministic blurring function $f_{\mathbf{B}}$ using $\tilde{\mathbf{x}}^\text{a}_0$.
The training and sampling processes are provided in Algorithms \ref{alg:opt}, \ref{alg:sample}.
\begin{table}
\vspace{-.3cm}
\begin{minipage}[t]{.47\textwidth}
\begin{algorithm}[H]
   \caption{Training}
   \label{alg:opt}
    \begin{algorithmic}
    \STATE Sample $\hat{\mathbf{x}}^\text{f} \sim p_{\text{RDKit}}(\mathbf{x}^\text{f})$
    \STATE Sample $\mathbf{x}_0^\text{a} \sim q(\mathbf{x}_0^\text{a})$
    \STATE Sample $t \sim \mathcal{U}[1, T]$
    \STATE Sample $\boldsymbol{\epsilon} \sim \mathcal{N}(\mathbf{0}, \sigma^2 \mathbf{I})$
    \State $\mathbf{x}_t^\text{a} \leftarrow f_{\mathbf{B}}(\mathbf{x}^\text{a}_0, \hat{\mathbf{x}}^\text{f}, t) + \boldsymbol{\epsilon}$
    \STATE Minimize $\|\mathbf{x}_0^\text{a} - \rho \big(f_\theta (\mathbf{x}_t^\text{a}, G, t) \big)\|^2$ 
    \end{algorithmic}
\end{algorithm}
\end{minipage}
\hfill
\begin{minipage}[t]{.47\textwidth}
\begin{algorithm}[H]
   \caption{Generation}
   \label{alg:sample}
    \begin{algorithmic}
    \STATE  Sample $\hat{\mathbf{x}}^\text{f} \sim p_{\text{RDKit}}(\mathbf{x}^\text{f})$
    \STATE $\mathbf{x}_T^\text{a} \leftarrow \mathbf{M} \hat{\mathbf{x}}^\text{f}$
    \FOR{$t$ in $\{T, \ldots, 1\}$}
    \STATE Sample $\boldsymbol{\epsilon} \sim \mathcal{N}(\mathbf{0}, \delta^2 \mathbf{I})$
    \STATE $\tilde{\mathbf{x}}_0^\text{a} \leftarrow f_\theta (\mathbf{x}_t^\text{a} + \boldsymbol{\epsilon}, \hat{\mathbf{x}}^\text{f}, G, t)$
    \STATE $\mathbf{x}_{t-1}^\text{a} \leftarrow f_{\mathbf{B}}(\tilde{\mathbf{x}}^\text{a}_0, \hat{\mathbf{x}}^\text{f}, t-1)$
    \ENDFOR
    \end{algorithmic}
\end{algorithm}
\end{minipage}
\vspace{-20pt}
\end{table}

\section{Related work}
\textbf{Hierarchical generation.} 
A hierarchical design of generative models is evident across multiple domains, including image generation \cite{menick2018generating, razavi2019generating, ho2022cascaded, rissanen2022generative} and speech synthesis \cite{hsu2018hierarchical, hono2020hierarchical}, aimed at enhancing the interpretability and quality of samples derived from coarse-grained information. In the field of computational biology, recent studies on molecular graph generation \cite{jin2018junction, jin2020hierarchical, geng2023de}, backmapping of protein structure \cite{yang2023chemically} and conformer generation \cite{wang2022generative} conditioned on the given ground truth coarse-grained information have reported the effectiveness of the hierarchical design. In recent unconditional conformer generation \cite{qiang2023coarse}, a denoising diffusion model \cite{hoogeboom2022equivariant} was exclusively used in the fragment structure generation step and not designed for coarse-to-fine generation.

% There exists an unconditional conformer generation method \cite{qiang2023coarse} that applied a denoising diffusion model into not on the coarse-to-fine generation but on the scaffold generation. the hierarchical framework. However, the diffusion model is confined to scaffold generation, while coarse-to-fine generation is achieved through four stages of iterative refinement.

\textbf{Data corruption in diffusion models.}
% Denoising diffusion models , which progressively introduce random noise to corrupt data and then generate samples from random noise, have shown remarkable generation ability across various data domains. 
The choice of data corruption can be considered a crucial aspect of the design space of diffusion models \cite{karras2022elucidating}, depending on the characteristics of the data domain and the specific problem definition. Recently, several studies on diffusion models have revealed that the choice of data corruption can be extended beyond random noise \cite{sohl2015deep, ho2020denoising, song2020denoising, nichol2021improved, rombach2022high} to methods such as masking \cite{bansal2023cold, Giannis2023soft, daras2023ambient}, blurring \cite{rissanen2022generative, bansal2023cold, Giannis2023soft, hoogeboom2022blurring}, and varying data dimension \cite{campbell2023trans}. We designed the data corruption process as a blurring operation in Euclidean space, transitioning from atom-level fine details to fragment-level coarse structures. This approach is more effective for multi-scale frameworks compared to random noise, which corrupts both fragment and atom geometries.

\section{Experiments} \label{sec:exp}
In this section, we evaluate our hierarchical molecular conformer generation framework via Equivariant Blurring Diffusion (EBD) on molecular conformer generation task. We conducted experiments to answer the following questions: i) \textbf{Ablation studies} (Sec. \ref{subsec:exp_ablation}): What are the effects of granularity of the fragment vocabulary, loss reparameterization, and data corruption processes of diffusion models? ii) \textbf{Geometric evaluation} (Sec. \ref{subsec:exp_geometric_eval}): Can EBD generate more diverse and accurate molecular conformers in Euclidean space than previous deep generative approaches? iii) \textbf{Property prediction} (Sec. \ref{subsec:exp_chemphy_eval}): Can EBD generate low-energy, stable conformers? 
% iv) \textbf{Generalization ability} (Sec. \ref{subsec:exp_generalization}): Can EBD generalize to out-of-distribution molecules?

% \begin{itemize}
%     \item \textbf{Ablation studies} [Sec. \ref{subsec:exp_ablation}]: What are the effects of granularity of the fragment vocabulary, loss reparameterization, and different trajectory of diffusion models?
%     \item \textbf{Geometric evaluation} [Sec. \ref{subsec:exp_geometric_eval}]: Can EBD generate more diverse and accurate molecular conformers in Euclidean space than previous deep generative approaches?
%     \item \textbf{Property prediction} [Sec. \ref{subsec:exp_chemphy_eval}]: Can EBD generate low-energy, stable conformers?
%     \item \textbf{Generalization ability} [Sec. \ref{subsec:exp_generalization}]: Can EBD generalize to out-of-distribution molecules?
% \end{itemize}

\subsection{Experiment setup} \label{subsec:exp_setup}
\textbf{Dataset.} We use GEOM-QM9 (QM9) \cite{ramakrishnan2014quantum} and GEOM-Drugs (Drugs) \cite{axelrod2022geom} which are small molecules and drug-like molecules, respectively. Each dataset comprises 40,000 molecules for the training set and 5,000 molecules for the validation set, with each molecule containing 5 conformers following data split of \cite{shi2021learning}. For the test set, we selected 200 molecules for each dataset, resulting in 22,408 and 14,324 conformers existing in QM9 and Drugs, respectively. The details of dataset were demonstrated in Appendix \ref{apdx:subsec_data}.

\textbf{Metrics.} To measure the accuracy and diversity of the generated conformer set $\mathcal{C}$, we adopted metrics proposed by \cite{ganea2021geomol}. The metrics are based on root-mean-square deviation (RMSD), which is a normalized Frobenius norm between two atomic coordinate matrices aligned using the Kabsch algorithm \cite{kabsch1976solution}. Given the ground truth conformer set $\mathcal{C}^*$ and the generated sample set $\mathcal{C}$, four metrics that follow precision and recall are defined as:
\begin{align}
    \text{COV-R} \ (\text{Recall}) &= \frac{1}{\vert \mathcal{C}^* \vert} \vert
    \{C^*\in \mathcal{C}^* \vert \text{RMSD}(C^*, C) \leq \delta,  C \in \mathcal{C} \} \vert, \label{eq:covr} \\
    \text{MAT-R} \ (\text{Recall}) &= \frac{1}{\vert \mathcal{C}^* \vert}
    \sum\limits_{C^* \in \mathcal{C}^*}
    \min\limits_{C \in \mathcal{C}} \text{RMSD}(C^*, C), \label{eq:matr}
\end{align}
where COV and MAT are coverage metric and matching metric \cite{xu2021learning}, respectively. COV quantifies the proportion of one set covered by another, with ``covered'' indicating RMSD values are within a threshold $\delta$. MAT measures the average of RMSD values of one conformer set with its closest conformer in another set. If $\mathcal{C}$ and $\mathcal{C}^*$ are exchanged in Eqs. (\ref{eq:covr}, \ref{eq:matr}), then metrics become COV-P (Precision) and MAT-P (Precision). The recall metric is focused on the diversity, while the precision metric measures the quality. The threshold $\delta$ is set to $0.5\text{\r{A}}$ for QM9 and $1.25\text{\r{A}}$ for Drugs. For each molecule, we generated conformers $C$ that are twice the size of the ground truth conformers $C^*$.

\textbf{Baselines.} We compare EBD to existing deep generative models for molecular conformer generation. The performance of RDKit \cite{landrum2013rdkit} that was used to generate the fragment structure of our model was measured as a baseline. Besides RDKit, machine learning models including CVGAE \cite{mansimov2019molecular}, GraphDG \cite{simm2020generative}, CGCF \cite{xu2021learning}, ConfVAE \cite{xu2021end},  GeoMol \cite{ganea2021geomol}, ConfGF \cite{shi2021learning}, and GeoDiff \cite{xu2022geodiff} were compared to our model. Among these, GeoDiff is the denoising diffusion model restoring the target distribution from random noise in atom coordinates. We adhered to their settings by configuring the maximum time step $T$ to 5,000. 
For EBD, we use the $T=50$, a noise scale of $0.01$ for the forward process ($\sigma$ in Eq. \ref{eq:for}) and $0.0125$ for the reverse process ($\delta$ in Eq. \ref{eq:rev}) in every experiments. The implementation details were reported in Appendix \ref{apdx:subsec_train}. 

\subsection{Ablation studies}\label{subsec:exp_ablation}
\begin{figure}
  \centering
  \includegraphics[trim={0 30cm 0 0}, clip, width=\textwidth]{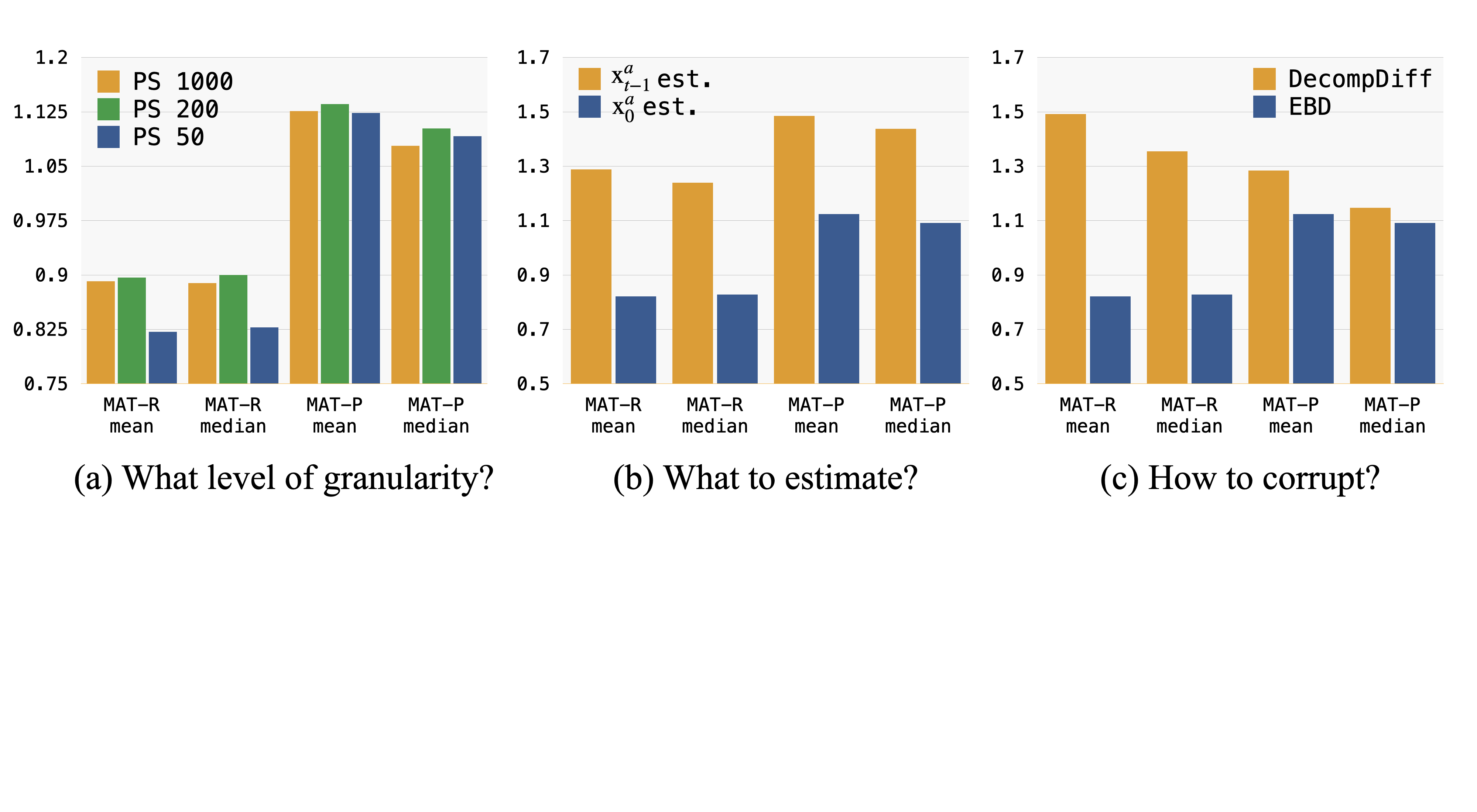}
  \caption{Ablation studies on the motivation and design choice of EBD. (a) Fragment vocabulary granularity; (b) Target of state estimator; (c) Choice of data corruption processes.}
\label{fig:ablation}
\vspace{-12pt}
\end{figure}

We conducted ablation studies to validate our model design, encompassing the size of the fragment vocabulary, the reparameterization of loss, and the blurring data corruption. For each ablation study, we calculated the mean and median of matching scores MAT-R and MAT-P on Drugs test set. Note that lower values of MAT-R and MAT-P indicate better results. 

% \begin{wraptable}[9]{R}{0.45\textwidth}
% \vspace{-20pt}
%     \centering
%     \caption{Statistics of fragments in GEOM-Drug.}
%     \label{tab:data}
%     \resizebox{\linewidth}{!}{
%     \begin{tabular}{l |cc}
%     \toprule
%     Vocab size &  #fragments/graph & #atoms/fragment
%     \midrule
%     50 & 11.77 & 4.02 \\
%     200 & 7.60 & 6.34 \\
%     1000 & 5.26 & 9.25 \\
%     \bottomrule
%     \end{tabular}
%     }
% \end{wraptable}

\begin{wraptable}{r}{0.35\textwidth}
\vspace{-12pt}
    \centering
    \caption{Statistics of fragment vocabulary $\mathcal{S}$ in Drugs.} 
    \vspace{-5pt}
    \label{tab:vocab_drug}
    \resizebox{\linewidth}{!}{
    \begin{tabular}{l|cc}
    \toprule  
    $\vert \mathcal{S} \vert$ &  $\# \text{frags} / G $ & $\# \text{atoms} / \text{frag} $ \\
    \midrule
        50 & 11.77 & 4.02 \\
        200 & 7.60 & 6.34 \\
        1000 & 5.26 & 9.25 \\
        \bottomrule
    \end{tabular}
    }
    \vspace{-10pt}
\end{wraptable} 

% \begin{wraptable}[9]{R}{0.55\textwidth}
% \vspace{-20pt}
%     \centering
%     \caption{MAE of predicted ensemble properties in eV.}
%     \label{tab:ablation_vocab}
%     % \vspace{-5pt}
%     % \scalebox{1}{
%     \resizebox{\linewidth}{!}{
%     \begin{tabular}{l | ccccc}
%     \toprule
%     Method & $ \overline{E} $ & $E_\text{min}$ & $ \overline{\Delta\epsilon} $ & $\Delta \epsilon_\text{min}$  & $\Delta \epsilon_\text{max}$ \\
%     \midrule
%     RDKit & 0.9233 & 0.6585 & 0.3698 & 0.8021 & 0.2359 \\
%     GraphDG & 9.1027 & 0.8882 & 1.7973 & 4.1743 & 0.4776 \\
%     CGCF & 28.9661 & 2.8410 & 2.8356 & 10.6361 & 0.5954 \\
%     ConfVAE & 8.2080 & 0.6100 & 1.6080 & 3.9111 & 0.2429 \\
%     ConfGF & 2.7886 & 0.1765 & 0.4688 & 2.1843 & \bf 0.1433 \\
%     GeoDiff &  0.25974 &  0.1551 &  0.3091 & 0.7033 & 0.1909 \\
%     EBD & \bf 0.18124 & \bf 0.1214 & \bf 0.1253 & \bf 0.5306 & 0.2153 \\
%     \bottomrule
%     \end{tabular}
%     }
% \end{wraptable}
\textbf{Effects of fragment granularity.} We assessed the performance variation as fragment structure became more detailed and informative by measuring the generation performances across different fragment vocabulary sizes $\vert \mathcal{S} \vert \in \{50, 200, 1000\}$. Since PS \cite{kong2022molecule}, the fragmentation method we used, initializes the vocabulary from unique single atoms, reducing the size $\vert \mathcal{S} \vert$ results in obtaining finer fragments $\hat{\mathbf{x}}^f$. We reported the statistics of the average number of fragments per graph ($\# \text{frags} / G $) and atoms per fragment ($\# \text{atoms} / \text{frag} $) in Table \ref{tab:vocab_drug} and the generation results in Fig. \ref{fig:ablation} (a). \begin{wraptable}{r}{0.55\textwidth}
\caption{Fine-to-fine generation on Drugs.}
  \vspace{-7pt}
  \label{tab:allatom}
  \centering
  \resizebox{0.55\textwidth}{!}{
    \begin{tabular}{l|cccc|cccc}
    \toprule[1.0pt]
    & \multicolumn{2}{c}{\shortstack[c]{COV-R ($\%$) $\uparrow$}}  & \multicolumn{2}{c|}{\shortstack[c]{MAT-R($\text{\r{A}}$) $\downarrow$}}  & \multicolumn{2}{c}{\shortstack[c]{COV-P ($\%$) $\uparrow$}}  & \multicolumn{2}{c}{\shortstack[c]{MAT-P ($\text{\r{A}}$) $\downarrow$}} \\
    % \cline{2-9}
     & Mean & Med & Mean & Med & Mean & Med & Mean & Med \\
    % \hline \hline
    \midrule[0.8pt]
    C2F  & \textbf{92.60}& \textbf{98.73} &0.8216 & 0.8279 & 66.24 & 68.39 & 1.1237 & 1.0916 \\ 
    F2F & 89.44& \textbf{98.73} &\textbf{0.7986} & \textbf{0.7710} & \textbf{76.62} & \textbf{88.64} & \textbf{1.0090} & \textbf{0.9397} \\ 
    \bottomrule[1.0pt]
    \end{tabular}
    }
\vspace{-10pt}
\end{wraptable}

 Thanks to the increased level of detail in fragments, $\vert \mathcal{S} \vert = 50$ can obtain better performance compared to other vocabulary sizes. This is because more specific fragment structures decrease the amount of atomic-level detail that needs to be generated. From this observation, we use $\vert \mathcal{S} \vert = 50$ in all subsequent experiments. We also conducted fine-to-fine generation to observe the ability of EBD. We measured the performance when the prior is $\hat{\mathbf{x}}^\text{a} \sim p_{\text{RDKit}}(\mathbf{x}^\text{a})$. Table \ref{tab:allatom} presents coarse-to-fine (C2F) result when $\vert \mathcal{S}\vert = 50$ and fine-to-fine (F2F) result. As expected, F2F shows more accurate results compared to C2F as it includes more detail in the prior distribution.

\textbf{Effects of loss reparameterization.} We presented the performance comparison between the less blurred previous state estimator in Eq. (\ref{eq:loss_prev}) and ground truth estimator in Eq. (\ref{eq:loss_gt}) after loss reprarameterization in Fig. \ref{fig:ablation} (b). From the previous state estimator, we acquired degenerated conformers with relatively high matching scores, which align with low FID score of IHDM \cite{rissanen2022generative} in image generation. On the other hand, we observed distinct advantages in introducing the ground truth estimator across all metrics. We speculate that the ground truth estimator facilitates the diffusion model in learning beyond locally blurring distributions towards the target distribution. 
\begin{wrapfigure}[19]{R}{0.5\textwidth}
\vspace{-10pt}
  \centering
  \includegraphics[trim={0 4cm 55cm 3cm}, clip, width=0.5\textwidth]{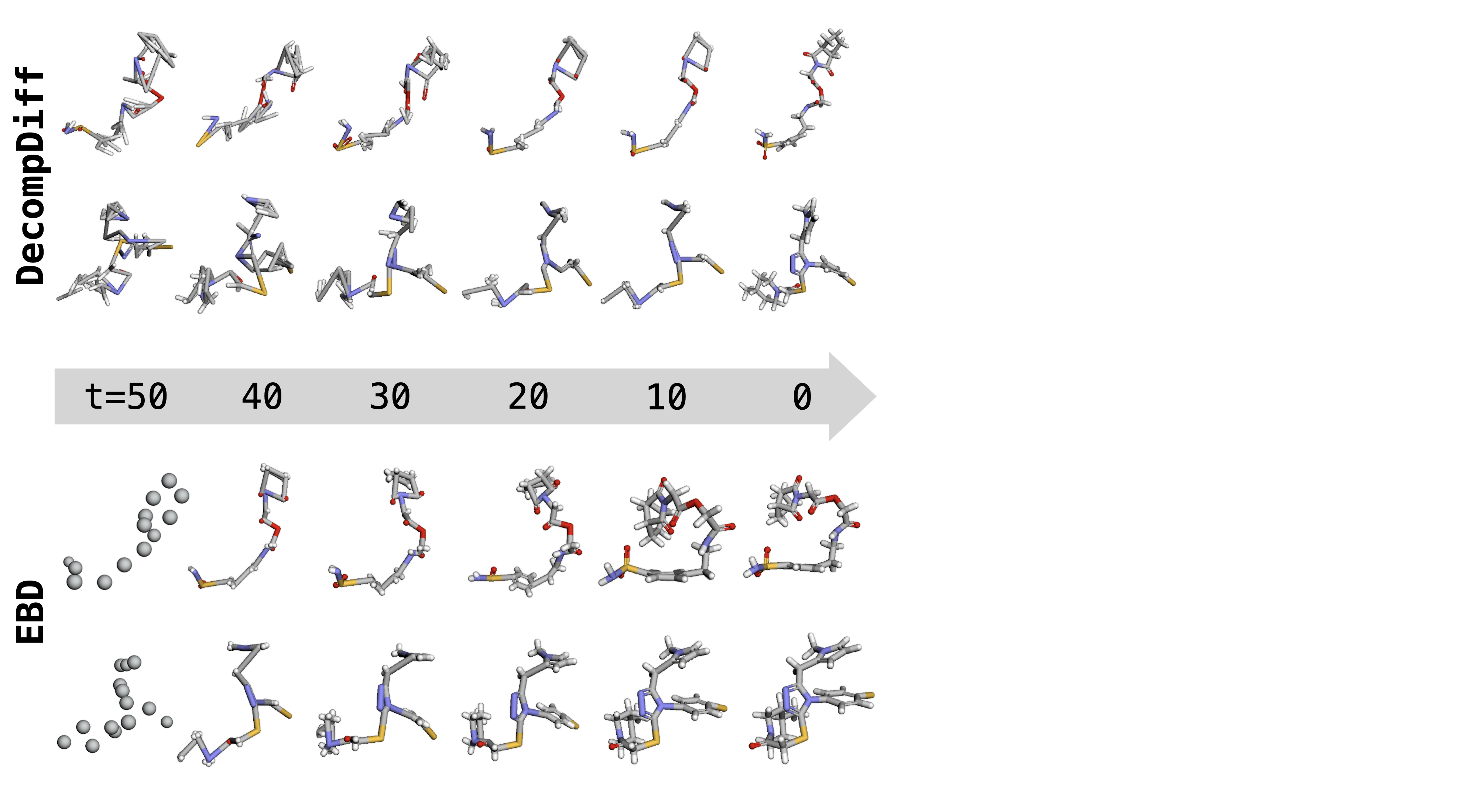}
  \caption{Sampling processes of two conformers depending on data corruptions.}
    \label{fig:traj}
\end{wrapfigure}

% \begin{figure}[h]
%   \centering
%   \includegraphics[trim={0 33cm 0 0}, clip, width=\textwidth]{Figures/trajectory_wider.png}
%   \caption{traj}
%   \label{fig:traj}
% \end{figure}

\textbf{Effects of data corruptions.} 
% DecompDiff \cite{guan2023decompdiff} is a denoising diffusion model that has the random noise prior distributions as the number of fragments in a molecule. 
We provide the same initial fragment structures $\hat{\mathbf{x}}^\text{f}$ to both EBD and DecompDiff \cite{guan2023decompdiff} so that the data corruption method becomes the primary distinction to examine.  DecompDiff is a better candidate than GeoDiff \cite{xu2022geodiff} to compare for this purpose because, unlike GeoDiff that generates conformers from a single prior distribution, DecompDiff denoises multiple prior distributions, where each mean corresponds to the coordinates of each fragment $\hat{\mathbf{x}}^\text{f}$. The generation results and sampling trajectories are compared between the two models ($T=50$ for both) in Fig. \ref{fig:ablation} (c) and Fig. \ref{fig:traj}. At first, we observed that the conformers generated from DecompDiff exhibit lower diversity scores compared to EBD. This is because the results of DecompDiff tend to adhere closely to the approximate fragment structure $\hat{\mathbf{x}}^\text{f}$, whereas EBD attempts to transition towards the ground truth fragment structure $\mathbf{x}^\text{f}$. We speculate that our blurring schedule, which entails a linear interpolation between $\mathbf{M}\hat{\mathbf{x}}^\text{f}$ and ${\mathbf{x}}^\text{a}_0$, facilitates the learning process for the diffusion model compared to a stochastic trajectory between prior and target distributions. As empirical evidence, we observed that DecompDiff primarily focuses on denoising the fragment structure throughout most of the sampling process in Fig. \ref{fig:traj}. On the other hand, EBD focuses on the entirety of the sampling process to generate fine details, resulting in better quality. 

% \begin{figure}
%   \centering
%   \includegraphics[trim={0 35.5cm 0 0}, clip, width=0.95\textwidth]{Figures/psd_x_ref.png}
%   \caption{Power spectral density analysis on forward processes of GeoDiff and EBD. The darkest line is the PSD coefficient of $\mathbf{x}^a_0$ and the lines become lighter as $t \rightarrow T$.}
%   \label{fig:psd}
%     \vspace{-13pt}
% \end{figure}

\begin{wrapfigure}[14]{R}{0.45\textwidth}
\vspace{-12pt}
     \centering
         \includegraphics[trim={0.4cm 0.7cm 0.4cm 0}, width=0.42\textwidth]{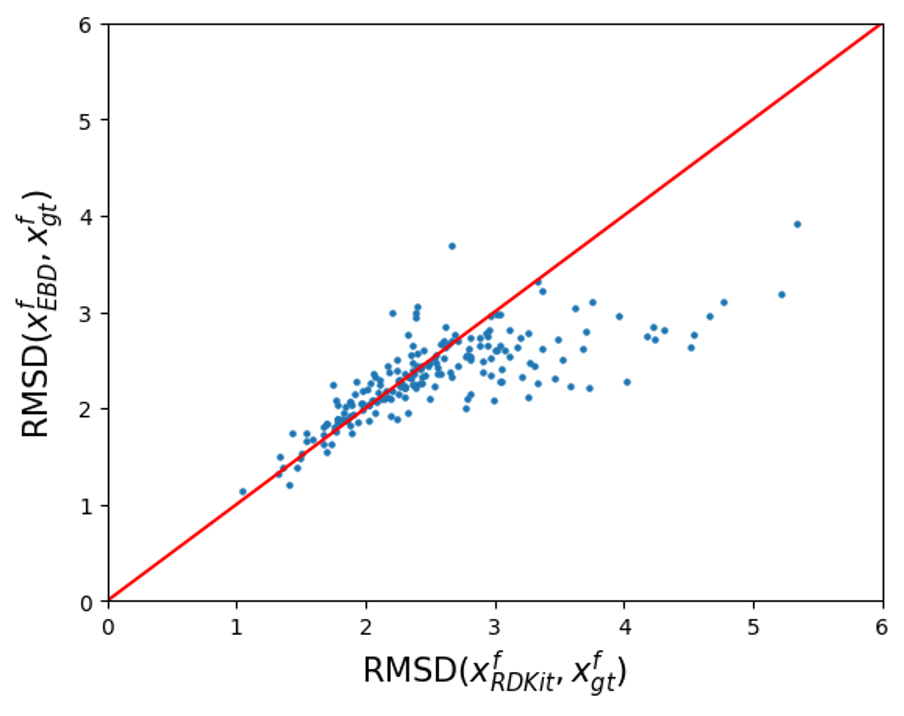}
        \caption{Correction on fragment coordinates}
        \label{fig:correction}
\end{wrapfigure} We conducted further analysis to observe how accurately the fragment coordinates generated by RDKit were corrected toward the ground truth by EBD. For 200 molecules in Drugs test set, we measured $\text{RMSD}(\mathbf{x}_\text{RDKit}^\text{f}, \mathbf{x}_\text{gt}^\text{f})$ and $\text{RMSD}(\mathbf{x}_\text{EBD}^\text{f}, \mathbf{x}_\text{gt}^\text{f})$. If $\text{RMSD}(\mathbf{x}_\text{EBD}^\text{f}, \mathbf{x}_\text{gt}^\text{f})$ is lower than $\text{RMSD}(\mathbf{x}_\text{RDKit}^\text{f}, \mathbf{x}_\text{gt}^\text{f})$ for a molecule, then it indicates that the model has corrected the fragment coordinates. In Fig. \ref{fig:correction}, the points below the red line represent cases where the model corrected the coordinates accurately. We observed that the greater the $\text{RMSD}(\mathbf{x}_\text{RDKit}^\text{f}, \mathbf{x}_\text{gt}^\text{f})$ (points further to the right on the x-axis), the larger the reduction in $\text{RMSD}(\mathbf{x}_\text{EBD}^\text{f}, \mathbf{x}_\text{gt}^\text{f})$. In other words, the lower the quality of the coarse-grained prior, the more accurately the model tends to make corrections. 

Additional visualizations of sampling results on Drugs and QM9, as well as the sampling processes, are illustrated in the Appendix \ref{apdx:vis}.

\subsection{Geometric evaluation}\label{subsec:exp_geometric_eval}
\begin{table}
  \caption{Geometric evaluation on Drugs ($\delta = 1.25\text{\r{A}}$).}
  \label{tab:rmsd_drug}
  \centering
  \resizebox{0.95\textwidth}{!}{
    \begin{tabular}{l|cccc|cccc}
    \toprule[1.0pt]
    & \multicolumn{2}{c}{\shortstack[c]{COV-R ($\%$) $\uparrow$}}  & \multicolumn{2}{c|}{\shortstack[c]{MAT-R($\text{\r{A}}$) $\downarrow$}}  & \multicolumn{2}{c}{\shortstack[c]{COV-P ($\%$) $\uparrow$}}  & \multicolumn{2}{c}{\shortstack[c]{MAT-P ($\text{\r{A}}$) $\downarrow$}} \\
    % \cline{2-9}
    Models & Mean & Med & Mean & Med & Mean & Med & Mean & Med \\
    % \hline \hline
    \midrule[0.8pt]
    RDKit & 45.74 & 31.75 & 1.5376 & 1.4004 & 54.78 & 59.48 & 1.3341 & 1.1996 \\ 
    CVGAE & 0.00 & 0.00 & 3.0702 & 2.9937 & - & - & - & - \\ 
    GraphDG & 8.27 & 0.00 & 1.9722 & 1.9845 & 2.08 & 0.00 & 2.4340 & 2.4100 \\ 
    CGCF & 53.96 & 57.06 & 1.2487 & 1.2247 & 21.68 & 13.72 & 1.8571 & 1.8066 \\
    ConfVAE & 55.20 & 59.43 & 1.2380 & 1.1417 & 22.96 & 14.05 & 1.8287 & 1.8159 \\
    GeoMol & 67.16 & 71.71 & 1.0875 & 1.0586 & - & - & - & - \\ 
    ConfGF & 62.15 & 70.93 & 1.1629 & 1.1596 & 23.42 & 15.52 & 1.7219 & 1.6863 \\
    GeoDiff {\scriptsize($T=5000$)} & 89.40 & 96.86 & 0.8571 & 0.8495 & 61.28 & 65.00 & 1.1642 & 1.1272 \\
    \midrule
    EBD {\scriptsize($T=50$)} & \textbf{92.60}& \textbf{98.73} &\textbf{0.8216} & \textbf{0.8279} & \textbf{66.24} & \textbf{68.39} & \textbf{1.1237} & \textbf{1.0916} \\ 
    \bottomrule[1.0pt]
    \end{tabular}
    }
% \vspace{-5pt}
\end{table}

We compared our hierarchical framework to the baseline RDKit and machine learning models for molecular conformer generation on Drugs, and the results are reported in Table \ref{tab:rmsd_drug}. EBD achieves superior performance across all metrics by generating diverse and accurate molecular conformers. In comparison to RDKit, which was used to generate fragment structures $\hat{\mathbf{x}}^\text{f}$, EBD achieved a significant improvement in the generation of diverse fine atomic details, as evidenced by higher COV-R and MAT-R scores. We also observed that, due to the informative fragment structure prior distribution and the proposed blurring schedule, EBD produces more diverse and higher-quality conformers with statistical significance (see Appendix~\ref{apdx:further_eval}), even with 100 times fewer $T$ compared to GeoDiff. We also reported EBD's better performance on QM9 with statistical significance in Appendix \ref{apdx:further_eval}.

% \begin{wrapfigure}[9]{R}{0.55\textwidth}
%   % \centering
%   \includegraphics[trim={0 2cm 70cm 0}, clip, width=0.5\textwidth]{Figures/psd.png}
%   \caption{Description of frameworks.}
% \end{wrapfigure}

\begin{figure}[h]
  \centering
  \includegraphics[trim={0 34cm 0 0}, clip, width=0.95\textwidth]{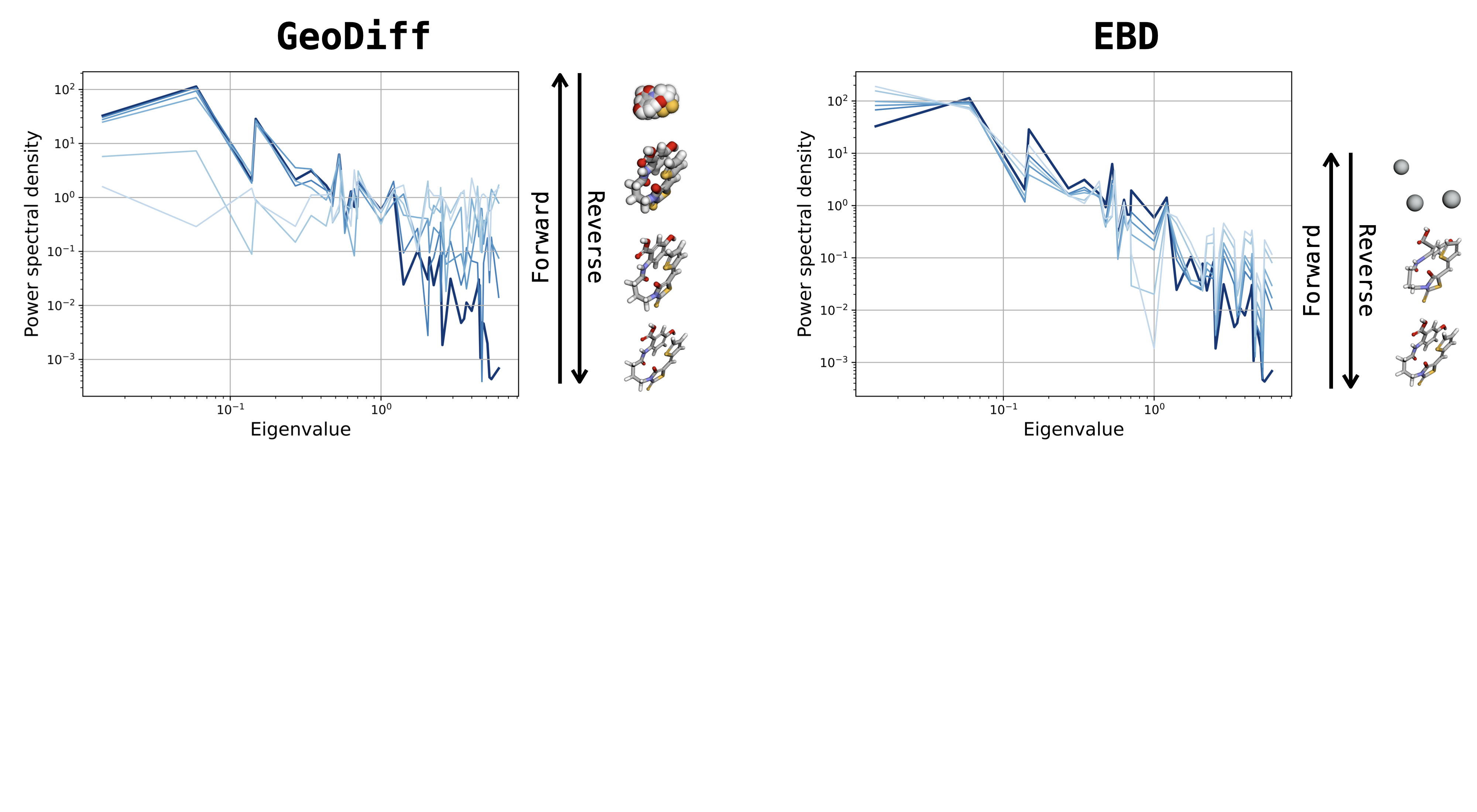}
  \caption{Power spectral density analysis on forward processes of GeoDiff and EBD. The darkest line is the PSD coefficient of $\mathbf{x}^\text{a}_0$ and the lines become lighter as $t \rightarrow T$.}
  \label{fig:psd}
    % \vspace{-5pt}
\end{figure}
We take an additional step to delve into the rationale behind the ability of EBD to achieve better performance with smaller $T$ by analyzing the spectral domain in Fig \ref{fig:psd}. Motivated from the analysis of \cite{rissanen2022generative}, we calculated the power spectral density, $\text{PSD}(\mathbf{x}^\text{a}_t)_{\mathtt{i}} = \frac{1}{3}\sum_{c \in \{x, y, z\}}\vert \mathbf{V}_{\mathtt{i}}^T \mathbf{x}^{\text{a}, c}_t \vert^2$, where $\mathbf{V}_{\mathtt{i}}$ is the $i$-th smallest eigenvector of graph Laplacian $\Delta$ of the molecular graph $G$, and $\mathbf{x}^{\text{a}, c}_t$ is one of the $\{x, y, z\}$ coordinate vectors of atomic coordinate matrix $\mathbf{x}^\text{a}_t$. The smaller eigenvalues correspond to coarser-grained structures, while higher eigenvalues correspond to finer details \cite{chung1997spectral}. Therefore, by measuring $\text{PSD}(\mathbf{x}^\text{a}_t)_{\mathtt{i}}$ during the forward processes $\{ \mathbf{x}^\text{a}_t\}_{t=0}^T$ of EBD and GeoDiff, we can ascertain which structural components are corrupted during the forward process and will be restored during the reverse process, respectively. For EBD, there is less significant perturbation across lower frequency parts of the PSD that corresponds to the coarser-grained discrepancy between ground truth $\mathbf{x}^\text{f}$ and approximate $\hat{\mathbf{x}}^\text{f}$ fragment positions. Thus, EBD primarily focuses on restoring perturbed fine-grained structures in the higher frequency parts throughout the entire generative process. This explains why EBD does not require excessive $T$. In contrast, GeoDiff requires relatively more $T$ because random noise corrupts the overall structural information, and the amount of perturbations is also significant.

\subsection{Chemical evaluation}\label{subsec:exp_chemphy_eval}
\begin{wraptable}[10]{R}{0.5\textwidth}
\vspace{-12pt}
    \centering
    \caption{Mean absolute errors between generated and ground truth ensemble properties in eV.}
    \vspace{-5pt}
    \label{tab:chem}
    % \vspace{-5pt}
    % \scalebox{1}{
    \resizebox{\linewidth}{!}{
    \begin{tabular}{l | ccccc}
    \toprule
    Models & $ \overline{E} $ & $E_\text{min}$ & $ \overline{\Delta\epsilon} $ & $\Delta \epsilon_\text{min}$  & $\Delta \epsilon_\text{max}$ \\
    \midrule
    RDKit & 0.9233 & 0.6585 & 0.3698 & 0.8021 & 0.2359 \\
    GraphDG & 9.1027 & 0.8882 & 1.7973 & 4.1743 & 0.4776 \\
    CGCF & 28.9661 & 2.8410 & 2.8356 & 10.6361 & 0.5954 \\
    ConfVAE & 8.2080 & 0.6100 & 1.6080 & 3.9111 & 0.2429 \\
    ConfGF & 2.7886 & 0.1765 & 0.4688 & 2.1843 & \bf 0.1433 \\
    GeoDiff &  0.2597 &  0.1551 &  0.3091 & 0.7033 & 0.1909 \\
    \midrule
    EBD & \bf 0.1812 & \bf 0.1214 & \bf 0.1253 & \bf 0.5306 & 0.2153 \\
    \bottomrule
    \end{tabular}
    }
\end{wraptable}
In addition to geometric evaluation, we assessed the quality of generated conformers by their chemical properties. After training EBD on QM9, following \cite{shi2021learning}, we generated 50 samples for each of the 30 molecules, which constitute a subset of QM9. Using PSI4 \cite{smith2020psi4}, we calculated properties of each conformer including the energy $\overline{E}$, the lowest energy $E_\text{min}$, HOMO-LUMO gap $\epsilon$, the average gap  $\overline{\Delta\epsilon}$, the minimum gap $\Delta \epsilon_\text{min}$ and the maximum gap $\Delta \epsilon_\text{max}$. Then, we measured the mean absolute errors between the properties of generated and ground truth (Table. \ref{tab:chem}). We observed that EBD can generate the most stable conformers compared to other methods, as evidenced by lower energy and HOMO-LUMO gap.

% \subsection{Generalization ability}\label{subsec:exp_generalization}
% % \input{Tables/rmsd_xl}
% Performance of GEOM-XL \cite{jing2022torsional}.
\section{Conclusion}
We introduced a novel hierarchical generative model for molecular conformers via Equivariant Blurring Diffusion (EBD), a diffusion model designed for coarse-to-fine generative scheme. After generating the initial distribution of fragment coordinates from a cheminformatics tool, EBD generated fine atomic details from coarse-grained structures through equivariant networks. We also proposed a simple and effective linear blurring scheduler and ground truth state estimator to enhance the model's ability to produce diverse and accurate conformers. Through extensive analysis of the proposed model and comparison between competitive denoising diffusion models, we substantiated the validity of the model design. We discussed a few limitations of our model in the Appendix \ref{apdx:limitation}.

\begin{ack}
We thank Yuning You and Hoseok Do for valuable feedback on the manuscript. This work was supported by NSF grant CCF-1943008. Portions of this research were conducted with the advanced computing resources provided by Texas A\&M High Performance Research Computing.
\end{ack}

\bibliographystyle{icml}
\bibliography{refs}

\begin{thebibliography}{62}
\providecommand{\natexlab}[1]{#1}
\providecommand{\url}[1]{\texttt{#1}}
\expandafter\ifx\csname urlstyle\endcsname\relax
  \providecommand{\doi}[1]{doi: #1}\else
  \providecommand{\doi}{doi: \begingroup \urlstyle{rm}\Url}\fi

\bibitem[Axelrod \& Gomez-Bombarelli(2022)Axelrod and
  Gomez-Bombarelli]{axelrod2022geom}
Axelrod, Simon and Gomez-Bombarelli, Rafael.
\newblock Geom, energy-annotated molecular conformations for property
  prediction and molecular generation.
\newblock \emph{Scientific Data}, 9\penalty0 (1):\penalty0 185, 2022.

\bibitem[Bansal et~al.(2023)Bansal, Borgnia, Chu, Li, Kazemi, Huang, Goldblum,
  Geiping, and Goldstein]{bansal2023cold}
Bansal, Arpit, Borgnia, Eitan, Chu, Hong-Min, Li, Jie, Kazemi, Hamid, Huang,
  Furong, Goldblum, Micah, Geiping, Jonas, and Goldstein, Tom.
\newblock Cold diffusion: Inverting arbitrary image transforms without noise.
\newblock \emph{Advances in Neural Information Processing Systems}, 36, 2023.

\bibitem[Campbell et~al.(2023)Campbell, Harvey, Weilbach, De~Bortoli,
  Rainforth, and Doucet]{campbell2023trans}
Campbell, Andrew, Harvey, William, Weilbach, Christian, De~Bortoli, Valentin,
  Rainforth, Thomas, and Doucet, Arnaud.
\newblock Trans-dimensional generative modeling via jump diffusion models.
\newblock \emph{Advances in Neural Information Processing Systems}, 36, 2023.

\bibitem[Chung(1997)]{chung1997spectral}
Chung, Fan~RK.
\newblock \emph{Spectral graph theory}, volume~92.
\newblock American Mathematical Soc., 1997.

\bibitem[Crippen et~al.(1988)Crippen, Havel, et~al.]{crippen1988distance}
Crippen, Gordon~M, Havel, Timothy~F, et~al.
\newblock \emph{Distance geometry and molecular conformation}, volume~74.
\newblock Research Studies Press Taunton, 1988.

\bibitem[Crouse(2016)]{crouse2016implementing}
Crouse, David~F.
\newblock On implementing 2d rectangular assignment algorithms.
\newblock \emph{IEEE Transactions on Aerospace and Electronic Systems},
  52\penalty0 (4):\penalty0 1679--1696, 2016.

\bibitem[Daras et~al.(2023{\natexlab{a}})Daras, Delbracio, Talebi, Dimakis, and
  Milanfar]{Giannis2023soft}
Daras, Giannis, Delbracio, Mauricio, Talebi, Hossein, Dimakis, Alexandros, and
  Milanfar, Peyman.
\newblock Soft diffusion: Score matching with general corruptions.
\newblock \emph{Transactions on Machine Learning Research}, 2023{\natexlab{a}}.

\bibitem[Daras et~al.(2023{\natexlab{b}})Daras, Shah, Dagan, Gollakota,
  Dimakis, and Klivans]{daras2023ambient}
Daras, Giannis, Shah, Kulin, Dagan, Yuval, Gollakota, Aravind, Dimakis, Alex,
  and Klivans, Adam.
\newblock Ambient diffusion: Learning clean distributions from corrupted data.
\newblock \emph{Advances in Neural Information Processing Systems}, 36,
  2023{\natexlab{b}}.

\bibitem[Degen et~al.(2008)Degen, Wegscheid-Gerlach, Zaliani, and
  Rarey]{degen2008art}
Degen, Jorg, Wegscheid-Gerlach, Christof, Zaliani, Andrea, and Rarey, Matthias.
\newblock On the art of compiling and using'drug-like'chemical fragment spaces.
\newblock \emph{ChemMedChem}, 3\penalty0 (10):\penalty0 1503, 2008.

\bibitem[Du et~al.(2022)Du, Zhang, Du, Meng, Chen, Zheng, Shao, and
  Liu]{du2022se}
Du, Weitao, Zhang, He, Du, Yuanqi, Meng, Qi, Chen, Wei, Zheng, Nanning, Shao,
  Bin, and Liu, Tie-Yan.
\newblock Se (3) equivariant graph neural networks with complete local frames.
\newblock In \emph{International Conference on Machine Learning}, pp.\
  5583--5608. PMLR, 2022.

\bibitem[Ganea et~al.(2021)Ganea, Pattanaik, Coley, Barzilay, Jensen, Green,
  and Jaakkola]{ganea2021geomol}
Ganea, Octavian, Pattanaik, Lagnajit, Coley, Connor, Barzilay, Regina, Jensen,
  Klavs, Green, William, and Jaakkola, Tommi.
\newblock Geomol: Torsional geometric generation of molecular 3d conformer
  ensembles.
\newblock \emph{Advances in Neural Information Processing Systems}, 34, 2021.

\bibitem[Geng et~al.(2023)Geng, Xie, Xia, Wu, Qin, Wang, Zhang, Wu, and
  Liu]{geng2023de}
Geng, Zijie, Xie, Shufang, Xia, Yingce, Wu, Lijun, Qin, Tao, Wang, Jie, Zhang,
  Yongdong, Wu, Feng, and Liu, Tie-Yan.
\newblock De novo molecular generation via connection-aware motif mining.
\newblock In \emph{The Eleventh International Conference on Learning
  Representations}, 2023.

\bibitem[Guan et~al.(2023)Guan, Zhou, Yang, Bao, Peng, Ma, Liu, Wang, and
  Gu]{guan2023decompdiff}
Guan, Jiaqi, Zhou, Xiangxin, Yang, Yuwei, Bao, Yu, Peng, Jian, Ma, Jianzhu,
  Liu, Qiang, Wang, Liang, and Gu, Quanquan.
\newblock Decompdiff: Diffusion models with decomposed priors for
  structure-based drug design.
\newblock In \emph{International Conference on Machine Learning}, pp.\
  11827--11846. PMLR, 2023.

\bibitem[Heusel et~al.(2017)Heusel, Ramsauer, Unterthiner, Nessler, and
  Hochreiter]{heusel2017gans}
Heusel, Martin, Ramsauer, Hubert, Unterthiner, Thomas, Nessler, Bernhard, and
  Hochreiter, Sepp.
\newblock Gans trained by a two time-scale update rule converge to a local nash
  equilibrium.
\newblock \emph{Advances in Neural Information Processing Systems}, 30, 2017.

\bibitem[Ho et~al.(2020)Ho, Jain, and Abbeel]{ho2020denoising}
Ho, Jonathan, Jain, Ajay, and Abbeel, Pieter.
\newblock Denoising diffusion probabilistic models.
\newblock \emph{Advances in Neural Information Processing Systems}, 33, 2020.

\bibitem[Ho et~al.(2022)Ho, Saharia, Chan, Fleet, Norouzi, and
  Salimans]{ho2022cascaded}
Ho, Jonathan, Saharia, Chitwan, Chan, William, Fleet, David~J, Norouzi,
  Mohammad, and Salimans, Tim.
\newblock Cascaded diffusion models for high fidelity image generation.
\newblock \emph{Journal of Machine Learning Research}, 23\penalty0
  (47):\penalty0 1--33, 2022.

\bibitem[Hono et~al.(2020)Hono, Tsuboi, Sawada, Hashimoto, Oura, Nankaku, and
  Tokuda]{hono2020hierarchical}
Hono, Yukiya, Tsuboi, Kazuna, Sawada, Kei, Hashimoto, Kei, Oura, Keiichiro,
  Nankaku, Yoshihiko, and Tokuda, Keiichi.
\newblock Hierarchical multi-grained generative model for expressive speech
  synthesis.
\newblock \emph{Interspeech 2020}, 2020.

\bibitem[Hoogeboom \& Salimans(2023)Hoogeboom and
  Salimans]{hoogeboom2022blurring}
Hoogeboom, Emiel and Salimans, Tim.
\newblock Blurring diffusion models.
\newblock In \emph{The Eleventh International Conference on Learning
  Representations}, 2023.

\bibitem[Hoogeboom et~al.(2022)Hoogeboom, Satorras, Vignac, and
  Welling]{hoogeboom2022equivariant}
Hoogeboom, Emiel, Satorras, V{\i}ctor~Garcia, Vignac, Cl{\'e}ment, and Welling,
  Max.
\newblock Equivariant diffusion for molecule generation in 3d.
\newblock In \emph{International Conference on Machine Learning}, pp.\
  8867--8887. PMLR, 2022.

\bibitem[Hsu et~al.(2019)Hsu, Zhang, Weiss, Zen, Wu, Cao, and
  Wang]{hsu2018hierarchical}
Hsu, Wei-Ning, Zhang, Yu, Weiss, Ron, Zen, Heiga, Wu, Yonghui, Cao, Yuan, and
  Wang, Yuxuan.
\newblock Hierarchical generative modeling for controllable speech synthesis.
\newblock In \emph{International Conference on Learning Representations}, 2019.

\bibitem[Jin et~al.(2018)Jin, Barzilay, and Jaakkola]{jin2018junction}
Jin, Wengong, Barzilay, Regina, and Jaakkola, Tommi.
\newblock Junction tree variational autoencoder for molecular graph generation.
\newblock In \emph{International Conference on Machine Learning}, pp.\
  2323--2332. PMLR, 2018.

\bibitem[Jin et~al.(2020)Jin, Barzilay, and Jaakkola]{jin2020hierarchical}
Jin, Wengong, Barzilay, Regina, and Jaakkola, Tommi.
\newblock Hierarchical generation of molecular graphs using structural motifs.
\newblock In \emph{International Conference on Machine Learning}, pp.\
  4839--4848. PMLR, 2020.

\bibitem[Jing et~al.(2022)Jing, Corso, Chang, Barzilay, and
  Jaakkola]{jing2022torsional}
Jing, Bowen, Corso, Gabriele, Chang, Jeffrey, Barzilay, Regina, and Jaakkola,
  Tommi.
\newblock Torsional diffusion for molecular conformer generation.
\newblock \emph{Advances in Neural Information Processing Systems}, 35, 2022.

\bibitem[Joshi et~al.(2023)Joshi, Bodnar, Mathis, Cohen, and
  Lio]{joshi2023expressive}
Joshi, Chaitanya~K, Bodnar, Cristian, Mathis, Simon~V, Cohen, Taco, and Lio,
  Pietro.
\newblock On the expressive power of geometric graph neural networks.
\newblock In \emph{International Conference on Machine Learning}, pp.\
  15330--15355. PMLR, 2023.

\bibitem[Kabsch(1976)]{kabsch1976solution}
Kabsch, Wolfgang.
\newblock A solution for the best rotation to relate two sets of vectors.
\newblock \emph{Acta Crystallographica Section A: Crystal Physics, Diffraction,
  Theoretical and General Crystallography}, 32\penalty0 (5):\penalty0 922--923,
  1976.

\bibitem[Karras et~al.(2022)Karras, Aittala, Aila, and
  Laine]{karras2022elucidating}
Karras, Tero, Aittala, Miika, Aila, Timo, and Laine, Samuli.
\newblock Elucidating the design space of diffusion-based generative models.
\newblock \emph{Advances in Neural Information Processing Systems}, 35, 2022.

\bibitem[Kingma et~al.(2021)Kingma, Salimans, Poole, and
  Ho]{kingma2021variational}
Kingma, Diederik, Salimans, Tim, Poole, Ben, and Ho, Jonathan.
\newblock Variational diffusion models.
\newblock \emph{Advances in Neural Information Processing Systems}, 34, 2021.

\bibitem[K{\"o}hler et~al.(2020)K{\"o}hler, Klein, and
  No{\'e}]{kohler2020equivariant}
K{\"o}hler, Jonas, Klein, Leon, and No{\'e}, Frank.
\newblock Equivariant flows: exact likelihood generative learning for symmetric
  densities.
\newblock In \emph{International Conference on Machine Learning}, pp.\
  5361--5370. PMLR, 2020.

\bibitem[Kong et~al.(2022)Kong, Huang, Tan, and Liu]{kong2022molecule}
Kong, Xiangzhe, Huang, Wenbing, Tan, Zhixing, and Liu, Yang.
\newblock Molecule generation by principal subgraph mining and assembling.
\newblock \emph{Advances in Neural Information Processing Systems}, 35, 2022.

\bibitem[Krizhevsky et~al.(2009)Krizhevsky, Hinton,
  et~al.]{krizhevsky2009learning}
Krizhevsky, Alex, Hinton, Geoffrey, et~al.
\newblock Learning multiple layers of features from tiny images.
\newblock 2009.

\bibitem[Landrum et~al.(2013)]{landrum2013rdkit}
Landrum, Greg et~al.
\newblock Rdkit: A software suite for cheminformatics, computational chemistry,
  and predictive modeling.
\newblock \emph{Greg Landrum}, 8\penalty0 (31.10):\penalty0 5281, 2013.

\bibitem[Lewell et~al.(1998)Lewell, Judd, Watson, and Hann]{lewell1998recap}
Lewell, Xiao~Qing, Judd, Duncan~B, Watson, Stephen~P, and Hann, Michael~M.
\newblock Recap retrosynthetic combinatorial analysis procedure: a powerful new
  technique for identifying privileged molecular fragments with useful
  applications in combinatorial chemistry.
\newblock \emph{Journal of chemical information and computer sciences},
  38\penalty0 (3):\penalty0 511--522, 1998.

\bibitem[Loshchilov \& Hutter(2019)Loshchilov and
  Hutter]{loshchilov2018decoupled}
Loshchilov, Ilya and Hutter, Frank.
\newblock Decoupled weight decay regularization.
\newblock In \emph{International Conference on Learning Representations}, 2019.

\bibitem[Mansimov et~al.(2019)Mansimov, Mahmood, Kang, and
  Cho]{mansimov2019molecular}
Mansimov, Elman, Mahmood, Omar, Kang, Seokho, and Cho, Kyunghyun.
\newblock Molecular geometry prediction using a deep generative graph neural
  network.
\newblock \emph{Scientific reports}, 9\penalty0 (1):\penalty0 20381, 2019.

\bibitem[Menick \& Kalchbrenner(2019)Menick and
  Kalchbrenner]{menick2018generating}
Menick, Jacob and Kalchbrenner, Nal.
\newblock Generating high fidelity images with subscale pixel networks and
  multidimensional upscaling.
\newblock In \emph{International Conference on Learning Representations}, 2019.

\bibitem[Nichol \& Dhariwal(2021)Nichol and Dhariwal]{nichol2021improved}
Nichol, Alexander~Quinn and Dhariwal, Prafulla.
\newblock Improved denoising diffusion probabilistic models.
\newblock In \emph{International Conference on Machine Learning}, pp.\
  8162--8171. PMLR, 2021.

\bibitem[Paszke et~al.(2017)Paszke, Gross, Chintala, Chanan, Yang, DeVito, Lin,
  Desmaison, Antiga, and Lerer]{paszke2017automatic}
Paszke, Adam, Gross, Sam, Chintala, Soumith, Chanan, Gregory, Yang, Edward,
  DeVito, Zachary, Lin, Zeming, Desmaison, Alban, Antiga, Luca, and Lerer,
  Adam.
\newblock Automatic differentiation in pytorch.
\newblock 2017.

\bibitem[Qiang et~al.(2023)Qiang, Song, Xu, Gong, Gao, Zhou, Ma, and
  Lan]{qiang2023coarse}
Qiang, Bo, Song, Yuxuan, Xu, Minkai, Gong, Jingjing, Gao, Bowen, Zhou, Hao, Ma,
  Wei-Ying, and Lan, Yanyan.
\newblock Coarse-to-fine: a hierarchical diffusion model for molecule
  generation in 3d.
\newblock In \emph{International Conference on Machine Learning}, pp.\
  28277--28299. PMLR, 2023.

\bibitem[Ramakrishnan et~al.(2014)Ramakrishnan, Dral, Rupp, and
  Von~Lilienfeld]{ramakrishnan2014quantum}
Ramakrishnan, Raghunathan, Dral, Pavlo~O, Rupp, Matthias, and Von~Lilienfeld,
  O~Anatole.
\newblock Quantum chemistry structures and properties of 134 kilo molecules.
\newblock \emph{Scientific data}, 1\penalty0 (1):\penalty0 1--7, 2014.

\bibitem[Razavi et~al.(2019)Razavi, Van~den Oord, and
  Vinyals]{razavi2019generating}
Razavi, Ali, Van~den Oord, Aaron, and Vinyals, Oriol.
\newblock Generating diverse high-fidelity images with vq-vae-2.
\newblock \emph{Advances in Neural Information Processing Systems}, 32, 2019.

\bibitem[Reidenbach \& Krishnapriyan(2023)Reidenbach and
  Krishnapriyan]{reidenbach2023coarsenconf}
Reidenbach, Danny and Krishnapriyan, Aditi.
\newblock Coarsenconf: Equivariant coarsening with aggregated attention for
  molecular conformer generation.
\newblock In \emph{NeurIPS 2023 Generative AI and Biology (GenBio) Workshop},
  2023.

\bibitem[Rissanen et~al.(2023)Rissanen, Heinonen, and
  Solin]{rissanen2022generative}
Rissanen, Severi, Heinonen, Markus, and Solin, Arno.
\newblock Generative modelling with inverse heat dissipation.
\newblock In \emph{The Eleventh International Conference on Learning
  Representations}, 2023.

\bibitem[Rombach et~al.(2022)Rombach, Blattmann, Lorenz, Esser, and
  Ommer]{rombach2022high}
Rombach, Robin, Blattmann, Andreas, Lorenz, Dominik, Esser, Patrick, and Ommer,
  Bj{\"o}rn.
\newblock High-resolution image synthesis with latent diffusion models.
\newblock In \emph{Proceedings of the IEEE/CVF Conference on Computer Vision
  and Pattern Recognition}, pp.\  10684--10695, 2022.

\bibitem[Saharia et~al.(2022)Saharia, Chan, Saxena, Li, Whang, Denton,
  Ghasemipour, Gontijo~Lopes, Karagol~Ayan, Salimans,
  et~al.]{saharia2022photorealistic}
Saharia, Chitwan, Chan, William, Saxena, Saurabh, Li, Lala, Whang, Jay, Denton,
  Emily~L, Ghasemipour, Kamyar, Gontijo~Lopes, Raphael, Karagol~Ayan, Burcu,
  Salimans, Tim, et~al.
\newblock Photorealistic text-to-image diffusion models with deep language
  understanding.
\newblock \emph{Advances in Neural Information Processing Systems}, 35, 2022.

\bibitem[Satorras et~al.(2021)Satorras, Hoogeboom, and Welling]{satorras2021n}
Satorras, V{\i}ctor~Garcia, Hoogeboom, Emiel, and Welling, Max.
\newblock E (n) equivariant graph neural networks.
\newblock In \emph{International Conference on Machine Learning}, pp.\
  9323--9332. PMLR, 2021.

\bibitem[Shi et~al.(2021)Shi, Luo, Xu, and Tang]{shi2021learning}
Shi, Chence, Luo, Shitong, Xu, Minkai, and Tang, Jian.
\newblock Learning gradient fields for molecular conformation generation.
\newblock In \emph{International Conference on Machine Learning}, pp.\
  9558--9568. PMLR, 2021.

\bibitem[Simm \& Hernandez-Lobato(2020)Simm and
  Hernandez-Lobato]{simm2020generative}
Simm, Gregor and Hernandez-Lobato, Jose~Miguel.
\newblock A generative model for molecular distance geometry.
\newblock In \emph{International Conference on Machine Learning}, pp.\
  8949--8958. PMLR, 2020.

\bibitem[Smith et~al.(2020)Smith, Burns, Simmonett, Parrish, Schieber,
  Galvelis, Kraus, Kruse, Di~Remigio, Alenaizan, et~al.]{smith2020psi4}
Smith, Daniel~GA, Burns, Lori~A, Simmonett, Andrew~C, Parrish, Robert~M,
  Schieber, Matthew~C, Galvelis, Raimondas, Kraus, Peter, Kruse, Holger,
  Di~Remigio, Roberto, Alenaizan, Asem, et~al.
\newblock Psi4 1.4: Open-source software for high-throughput quantum chemistry.
\newblock \emph{The Journal of chemical physics}, 152\penalty0 (18), 2020.

\bibitem[Sohl-Dickstein et~al.(2015)Sohl-Dickstein, Weiss, Maheswaranathan, and
  Ganguli]{sohl2015deep}
Sohl-Dickstein, Jascha, Weiss, Eric, Maheswaranathan, Niru, and Ganguli, Surya.
\newblock Deep unsupervised learning using nonequilibrium thermodynamics.
\newblock In \emph{International Conference on Machine Learning}, pp.\
  2256--2265. PMLR, 2015.

\bibitem[Song et~al.(2020{\natexlab{a}})Song, Meng, and
  Ermon]{song2020denoising}
Song, Jiaming, Meng, Chenlin, and Ermon, Stefano.
\newblock Denoising diffusion implicit models.
\newblock In \emph{International Conference on Learning Representations},
  2020{\natexlab{a}}.

\bibitem[Song \& Ermon(2019)Song and Ermon]{song2019generative}
Song, Yang and Ermon, Stefano.
\newblock Generative modeling by estimating gradients of the data distribution.
\newblock \emph{Advances in Neural Information Processing Systems}, 32, 2019.

\bibitem[Song et~al.(2020{\natexlab{b}})Song, Sohl-Dickstein, Kingma, Kumar,
  Ermon, and Poole]{song2020score}
Song, Yang, Sohl-Dickstein, Jascha, Kingma, Diederik~P, Kumar, Abhishek, Ermon,
  Stefano, and Poole, Ben.
\newblock Score-based generative modeling through stochastic differential
  equations.
\newblock In \emph{International Conference on Learning Representations},
  2020{\natexlab{b}}.

\bibitem[Thomas et~al.(2018)Thomas, Smidt, Kearnes, Yang, Li, Kohlhoff, and
  Riley]{thomas2018tensor}
Thomas, Nathaniel, Smidt, Tess, Kearnes, Steven, Yang, Lusann, Li, Li,
  Kohlhoff, Kai, and Riley, Patrick.
\newblock Tensor field networks: Rotation-and translation-equivariant neural
  networks for 3d point clouds.
\newblock \emph{arXiv preprint arXiv:1802.08219}, 2018.

\bibitem[Vahdat et~al.(2021)Vahdat, Kreis, and Kautz]{vahdat2021score}
Vahdat, Arash, Kreis, Karsten, and Kautz, Jan.
\newblock Score-based generative modeling in latent space.
\newblock \emph{Advances in Neural Information Processing Systems}, 34, 2021.

\bibitem[Vahdat et~al.(2022)Vahdat, Williams, Gojcic, Litany, Fidler, Kreis,
  et~al.]{vahdat2022lion}
Vahdat, Arash, Williams, Francis, Gojcic, Zan, Litany, Or, Fidler, Sanja,
  Kreis, Karsten, et~al.
\newblock Lion: Latent point diffusion models for 3d shape generation.
\newblock \emph{Advances in Neural Information Processing Systems},
  35:\penalty0 10021--10039, 2022.

\bibitem[Wang et~al.(2022)Wang, Xu, Cai, Miller, Smidt, Wang, Tang, and
  Gomez-Bombarelli]{wang2022generative}
Wang, Wujie, Xu, Minkai, Cai, Chen, Miller, Benjamin~K, Smidt, Tess, Wang,
  Yusu, Tang, Jian, and Gomez-Bombarelli, Rafael.
\newblock Generative coarse-graining of molecular conformations.
\newblock In \emph{International Conference on Machine Learning}, pp.\
  23213--23236. PMLR, 2022.

\bibitem[Xu et~al.(2021{\natexlab{a}})Xu, Luo, Bengio, Peng, and
  Tang]{xu2021learning}
Xu, Minkai, Luo, Shitong, Bengio, Yoshua, Peng, Jian, and Tang, Jian.
\newblock Learning neural generative dynamics for molecular conformation
  generation.
\newblock In \emph{International Conference on Learning Representations},
  2021{\natexlab{a}}.

\bibitem[Xu et~al.(2021{\natexlab{b}})Xu, Wang, Luo, Shi, Bengio,
  Gomez-Bombarelli, and Tang]{xu2021end}
Xu, Minkai, Wang, Wujie, Luo, Shitong, Shi, Chence, Bengio, Yoshua,
  Gomez-Bombarelli, Rafael, and Tang, Jian.
\newblock An end-to-end framework for molecular conformation generation via
  bilevel programming.
\newblock In \emph{International Conference on Machine Learning}, pp.\
  11537--11547. PMLR, 2021{\natexlab{b}}.

\bibitem[Xu et~al.(2022)Xu, Yu, Song, Shi, Ermon, and Tang]{xu2022geodiff}
Xu, Minkai, Yu, Lantao, Song, Yang, Shi, Chence, Ermon, Stefano, and Tang,
  Jian.
\newblock Geodiff: A geometric diffusion model for molecular conformation
  generation.
\newblock In \emph{International Conference on Learning Representations}, 2022.

\bibitem[Xu et~al.(2023)Xu, Powers, Dror, Ermon, and Leskovec]{xu2023geometric}
Xu, Minkai, Powers, Alexander~S, Dror, Ron~O, Ermon, Stefano, and Leskovec,
  Jure.
\newblock Geometric latent diffusion models for 3d molecule generation.
\newblock In \emph{International Conference on Machine Learning}, pp.\
  38592--38610. PMLR, 2023.

\bibitem[Yang \& G{\'o}mez-Bombarelli(2023)Yang and
  G{\'o}mez-Bombarelli]{yang2023chemically}
Yang, Soojung and G{\'o}mez-Bombarelli, Rafael.
\newblock Chemically transferable generative backmapping of coarse-grained
  proteins.
\newblock In \emph{International Conference on Machine Learning}, pp.\
  39277--39298, 2023.

\bibitem[Zhu et~al.(2022)Zhu, Xia, Liu, Wu, Xie, Wang, Wang, Qin, Zhou, Li,
  et~al.]{zhu2022direct}
Zhu, Jinhua, Xia, Yingce, Liu, Chang, Wu, Lijun, Xie, Shufang, Wang, Yusong,
  Wang, Tong, Qin, Tao, Zhou, Wengang, Li, Houqiang, et~al.
\newblock Direct molecular conformation generation.
\newblock \emph{Transactions on Machine Learning Research}, 2022.

\end{thebibliography}

\newpage
\appendix
\renewcommand{\theequation}{A.\arabic{equation}}
\setcounter{equation}{0}

% Optionally include supplemental material (complete proofs, additional experiments and plots) in appendix.
% All such materials \textbf{SHOULD be included in the main submission.}

\section{Deblurring network architectures} \label{apdx:networks}
In SE(3)-equivariant deblurring networks, there are update functions of SE(3)-invariant fragment and atom features $\mathbf{h}^\text{f}$, $\mathbf{h}^\text{a}$, as well as an update function of SE(3)-equivariant atom coordinates $\mathbf{x}^\text{a}$ motivated from equivariant graph neural networks \cite{satorras2021n}.
For the fragments, we constructed a complete graph to account for dense interactions among them. In the case of atoms, we expanded the neighbor set of each atom by including multi-hop neighbors derived from the powers of the adjacency matrix and a radius graph, which includes atoms within a specified cutoff distance. The benefits of dense interactions for accurate conformers estimation have been confirmed in several studies \cite{simm2020generative, xu2022geodiff, hoogeboom2022equivariant}.

The architecture of the SE(3)-invariant message passing and feature update functions at the fragment- and atom-level is as follows:
\begin{align} 
    \mathbf{m}^\text{f}_{\mathtt{ij}} &= \phi_m^\text{f} (\mathbf{h}_{\mathtt{i}}^{\text{f},l}, \mathbf{h}_{\mathtt{j}}^{\text{f},l}, \|\mathbf{x}^\text{f}_{\mathtt{i}} - \mathbf{x}^\text{f}_{\mathtt{j}}\|), & \mathbf{h}_{\mathtt{i}}^{\text{f},l+1} &= \phi_h^\text{f} (\mathbf{h}_{\mathtt{i}}^{\text{f},l}, \sum\nolimits_{\mathtt{{j}} \in N(\mathbf{x}_{\mathtt{i}}^\text{f})} \mathbf{m}_{\mathtt{ij}}^\text{f}, \mathbf{h}^{\text{a},l}), \\
    \mathbf{m}^\text{a}_{\mathtt{ij}} &= \phi_m^\text{a} (\mathbf{h}_{\mathtt{i}}^{\text{a},l}, \mathbf{h}_{\mathtt{j}}^{\text{a},l}, \|\mathbf{x}_{\mathtt{i}}^{\text{a},l} - \mathbf{x}_{\mathtt{j}}^{\text{a},l}\|, e^\text{a}_{\mathtt{ij}}), & \mathbf{h}_{\mathtt{i}}^{\text{a},l+1} &= \phi_h^\text{a} (\mathbf{h}_{\mathtt{i}}^{\text{a},l}, \sum\nolimits_{\mathtt{j} \in N(\mathbf{x}_{\mathtt{i}}^\text{a})} \mathbf{m}_{\mathtt{ij}}^\text{a}, \mathbf{h}^{\text{f},l+1}),
\end{align} 
where $\mathbf{m}_{\mathtt{ij}} \in \mathbb{R}^{d}$ is the message for each interactions, and $\mathbf{h} \in \mathbb{R}^{d}$ is the feature vector from the aggregated messages and features from different hierarchy level. For every invariant update functions $\phi_m^\text{f}, \phi_h^\text{f}, \phi_m^\text{a}, \phi_h^\text{a}$, we used multilayer perceptrons. For initial features $\mathbf{h}_{\mathtt{i}}^{\text{f}, 0}$ of fragments, we defined a $3$-dimensional vector as a frequency histogram of its constituent atom types based on their chemical properties, including hydrophobicity, hydrogen bond center, and negative charge center following \cite{qiang2023coarse}. The detailed definition of the initial fragment features is in Table \ref{tab:frag_feat}. For initial atom features $\mathbf{h}_{\mathtt{i}}^{\text{a}, 0} \in \mathbb{R}^{d}$ and bond features $e^\text{a}_{\mathtt{ij}} \in \mathbb{R}^{d}$, we used embeddings from atom types and bond types, respectively.

\begin{table}[h]
\vspace{-12pt}
  \caption{Initial fragment feature based on chemical properties}
  \label{tab:frag_feat}
  \centering
  \begin{tabular}{lll}
    \toprule
    Properties     & Details     & Types\\
    \midrule
    Hydrophobicity           & Frequency of C element  & Integer    \\
    Hydrogen bond center     & Frequency of O, N, S, P elements & Integer    \\
    Negative charge center   & Frequency of F, Cl, Br, I elements       & Integer  \\
    \bottomrule
  \end{tabular}
\end{table}
For the $i$-th atom $\mathbf{x}_{\mathtt{i}}^\text{a}$ belongs to the $k$-th fragment $\mathbf{x}_{\mathtt{k}}^\text{f}$, the architecture of the equivariant atom coordinate update function is as follows:
\begin{align}
\begin{split} \label{eq:apdx_equi}
\mathbf{x}_{\mathtt{i}}^{\text{a},l+1} = \mathbf{x}_{\mathtt{i}}^{\text{a},l} &+ \sum_{\mathtt{j} \in N(\mathbf{x}_{\mathtt{i}}^\text{a})} \frac{\mathbf{x}_{\mathtt{i}}^{\text{a},l} - \mathbf{x}_{\mathtt{j}}^{\text{a},l}}{d_{\mathtt{ij}}^{\text{a},l} + 1} \phi_x^\text{a}( \mathbf{h}_{\mathtt{i}}^{\text{a},l+1}, \mathbf{h}_{\mathtt{j}}^{\text{a},l+1}, \mathbf{m}^\text{a}_{\mathtt{ij}},  e^\text{a}_{\mathtt{ij}}) \\ 
&+ \frac{\mathbf{x}_{\mathtt{i}}^{\text{a},l} - \mathbf{x}_{\mathtt{k}}^\text{f}}{\|\mathbf{x}_{\mathtt{i}}^{\text{a},l} - \mathbf{x}_{\mathtt{k}}^\text{f}\| + 1} \phi_x^\text{f}(\mathbf{h}_{\mathtt{i}}^{\text{a},l+1},  \mathbf{h}_{\mathtt{k}}^{\text{f},l+1}, \|\mathbf{x}_{\mathtt{i}}^{\text{a},l} - \mathbf{x}_{\mathtt{k}}^\text{f}\|),
\end{split}
\end{align}
where $\mathbf{x}^\text{f}_{\mathtt{k}}$ is the $k$-th row of $\mathbf{M}^\dagger \mathbf{x}^\text{a}_t$, and $d_{\mathtt{ij}}^{\text{a}, l} = \|\mathbf{x}_{\mathtt{i}}^{\text{a}, l} - \mathbf{x}_{\mathtt{j}}^{\text{a}, l}\|$ are inter-atomic distances. For every equivariant update functions $\phi_x^\text{a}, \phi_x^\text{f}$, we used multilayer perceptrons. For three terms in right-hand side of Eq. (\ref{eq:apdx_equi}), the first term is the coordinate from the previous layer, the second term is an equivariant update function that accounts for atom-level interactions, and the third term is an equivariant update function that considers the deviation of the current atom coordinate from its respective fragment's coordinate.

\section{Derivation of loss function} \label{apdx:loss}
In this section, we explain the derivation of the loss function for the ground truth state estimator from the previous state estimator. The loss function of previous state estimation is defined as:
\begin{equation} 
    L_{t-1} = \mathbb{E}_{t, \mathbf{x}^\text{a}_0, \mathbf{x}^\text{a}_t, \hat{\mathbf{x}}^\text{f} } [\|  f_{\mathbf{B}}(\mathbf{x}^\text{a}_0, \hat{\mathbf{x}}^\text{f}, t-1) - \rho \big(\mu_{\theta}(\mathbf{x}^\text{a}_t, \hat{\mathbf{x}}^\text{f}, G, t)\big) \|^2],
\end{equation}
where $\rho$ is the Kabsch algorithm \cite{kabsch1976solution} to obtain the optimal rotation matrix for alignment. Through alignment $\rho$ of the prediction from $\mu_\theta$ to the less blurred state $f_{\mathbf{B}}(\mathbf{x}^\text{a}_0, \hat{\mathbf{x}}^\text{f}, t-1)$ after translating both terms to the zero center-of-mass subspace, the loss function becomes invariant to the translation and rotation of the prediction. 

However, this previous state estimator generates unsatisfactory conformers as empirically observed in Sec. \ref{subsec:exp_ablation}. We conjectured the reason as the model limited to learn the locally small steps towards the ground truth distribution at each time step \cite{Giannis2023soft}.
% and possibly suffering from the lever-arm effect of error propagation. 
Thus, we reparameterize $\mu_{\theta}(\mathbf{x}^\text{a}_t, \hat{\mathbf{x}}^\text{f}, G, t)$ as $(1 - \frac{t-1}{T}) f_{\theta}(\mathbf{x}^\text{a}_t, G, t) + \frac{t-1}{T} \mathbf{M}\hat{\mathbf{x}}^\text{f}$ to make the deblurring network estimates the ground truth state $\mathbf{x}^\text{a}_0$ instead of the previous less blurred state via neural networks $f_{\theta}$. We first start with the non-invariant previous state estimation, which is without the alignment $\rho$:
\begin{align}
    L_{t-1} &= \mathbb{E}_{t, \mathbf{x}^\text{a}_0, \mathbf{x}^\text{a}_t, \hat{\mathbf{x}}^\text{f} } [\|  f_{\mathbf{B}}(\mathbf{x}^\text{a}_0, \hat{\mathbf{x}}^\text{f}, t-1) - \mu_{\theta}(\mathbf{x}^\text{a}_t, \hat{\mathbf{x}}^\text{f}, G, t) \|^2], \\
    &= \mathbb{E}_{t, \mathbf{x}^\text{a}_0, \mathbf{x}^\text{a}_t, \hat{\mathbf{x}}^\text{f}} [\|  f_{\mathbf{B}}(\mathbf{x}^\text{a}_0, \hat{\mathbf{x}}^\text{f}, t-1) - (1 - \tfrac{t-1}{T}) f_{\theta}(\mathbf{x}^\text{a}_t, G, t) - \tfrac{t-1}{T} \mathbf{M}\hat{\mathbf{x}}^\text{f} \|^2] \\
    &= \mathbb{E}_{t, \mathbf{x}^\text{a}_0, \mathbf{x}^\text{a}_t, \hat{\mathbf{x}}^\text{f}} [\|  (1 - \tfrac{t-1}{T}) \mathbf{x}^\text{a}_0 + \tfrac{t-1}{T} \mathbf{M}\hat{\mathbf{x}}^\text{f} - (1 - \tfrac{t-1}{T}) f_{\theta}(\mathbf{x}^\text{a}_t, G, t) - \tfrac{t-1}{T} \mathbf{M}\hat{\mathbf{x}}^\text{f}  \|^2] \\
    &= \mathbb{E}_{t, \mathbf{x}^\text{a}_0, \mathbf{x}^\text{a}_t, \hat{\mathbf{x}}^\text{f}} [\|  (1 - \tfrac{t-1}{T}) (\mathbf{x}^\text{a}_0 - f_{\theta}(\mathbf{x}^\text{a}_t, G, t)) + \tfrac{t-1}{T}(\mathbf{M}\hat{\mathbf{x}}^\text{f} - \mathbf{M}\hat{\mathbf{x}}^\text{f} ) \|^2] \\
    &= \mathbb{E}_{t, \mathbf{x}^\text{a}_0, \mathbf{x}^\text{a}_t, \hat{\mathbf{x}}^\text{f}} (1 - \tfrac{t-1}{T})^2 [\| \mathbf{x}^\text{a}_0 - f_{\theta}(\mathbf{x}^\text{a}_t, G, t) \|^2] \label{apdx:eq_original_loss} \\
    & \approx \mathbb{E}_{t, \mathbf{x}^\text{a}_0, \mathbf{x}^\text{a}_t, \hat{\mathbf{x}}^\text{f}} [\| \mathbf{x}^\text{a}_0 - \rho \big( f_{\theta}(\mathbf{x}^\text{a}_t, G, t) \big) \|^2] \label{apdx:eq_final_loss}.
\end{align}
In the last stage from  Eq. (\ref{apdx:eq_original_loss}) to Eq. (\ref{apdx:eq_final_loss}), we simplified the loss function by discarding the time-dependent weight as \cite{ho2020denoising}. Finally, we make the loss function for ground truth estimation invariant by aligning the prediction from $f_\theta$ to the ground truth state using Kabsch alignment $\rho$ \cite{kabsch1976solution}.

\section{Implementation details} \label{apdx:implementation}
\subsection{Datasets} \label{apdx:subsec_data}
We used GEOM-QM9 (QM9) \cite{ramakrishnan2014quantum} and GEOM-Drugs (Drugs) \cite{axelrod2022geom} for analysis and comparison between molecular conformer generation models. Each dataset comprises 40,000 molecules for the training set and 5,000 molecules for the validation set, with each molecule containing 5 conformers following data split of \cite{shi2021learning}. We obtained the raw data, the pre-processed data and the data split at \url{https://github.com/DeepGraphLearning/ConfGF}. For the test set, we selected 200 molecules for each dataset, resulting in 22,408 and 14,324 conformers existing in QM9 and Drugs, respectively.

For fragmentation of the molecular graphs $G = (\mathcal{V}, \mathcal{E})$ in Drugs and QM9, we used Principal Subgraph (PS) \cite{kong2022molecule} (\url{https://github.com/THUNLP-MT/PS-VAE}) which can construct a fragment vocabulary $\mathcal{S}$ whose elements are the largest and frequent repetitive subgraphs of molecules. Starting from all unique atoms in $\mathcal{S}$ at initial stage, PS iteratively merges neighboring fragments. The most frequent fragment among the newly merged fragments was added to the vocabulary at each iteration, and this operation was repeated until the desired size of the vocabulary was reached. Thus, the smaller the fragment vocabulary, the finer fragments can be obtained. One of the advantages of PS compared to existing fragmentation methods such as RECAP \cite{lewell1998recap}, BRICS \cite{degen2008art}, junction tree decomposition \cite{jin2018junction} is the ability to control the vocabulary size, allowing us to observe how performance varies with fragment granularity. We constructed $\mathcal{S}$ for each dataset with three fragment vocabulary sizes $\vert \mathcal{S} \vert \in \{50, 200, 1000\}$. The average numbers of fragments per graph ($\# \text{frags} / G $) and atoms per fragment ($\# \text{atoms} / \text{frag} $) of Drugs and QM9 were reported in Table \ref{tab:vocab}.
% \begin{wraptable}{r}{0.33\textwidth}
% \vspace{-10pt}
%     \centering
%     \caption{Statistics of fragment vocabulary $\mathcal{S}$ in GEOM-QM9.} 
%     \label{tab:vocab_qm9}
%     \resizebox{\linewidth}{!}{
%     \begin{tabular}{l|cc}
%     \toprule  
%     & \multicolumn{2}{c}{QM9}   & \multicolumn{2}{c}{Drugs}      
%     $\vert \mathcal{S} \vert$ &  $\# \text{frags} / G $ & $\# \text{atoms} / \text{frag} $ & $\# \text{frags} / G $ & $\# \text{atoms} / \text{frag} $ \\
%     \midrule
%         50 & 5.17 & 3.91 \\
%         200 & 3.70 & 5.45 \\
%         1000 & 2.91 & 6.98 \\
%         \bottomrule
%     \end{tabular}
%     }
%     \vspace{-5pt}
% \end{wraptable} 

\begin{table}[ht]
\vspace{-12pt}
    \centering
    \caption{Statistics of fragment vocabulary $\mathcal{S}$.} 
    \label{tab:vocab}
    \begin{tabular}{l|cc|cc}
    \toprule  
    & \multicolumn{2}{c}{Drugs}   & \multicolumn{2}{|c}{QM9}    \\  
    $\vert \mathcal{S} \vert$ &  $\# \text{frags} / G $ & $\# \text{atoms} / \text{frag} $ & $\# \text{frags} / G $ & $\# \text{atoms} / \text{frag} $ \\
    \midrule
        50 & 11.77 & 4.02 & 5.17 & 3.91  \\
        200 & 7.60 & 6.34 & 3.70 & 5.45  \\
        1000 & 5.26 & 9.25 & 2.91 & 6.98 \\
        \bottomrule
  \end{tabular}
\end{table}

\begin{figure}[t!]
  \centering
  \includegraphics[trim={0 30.5cm 25cm 0}, clip, width=\textwidth]{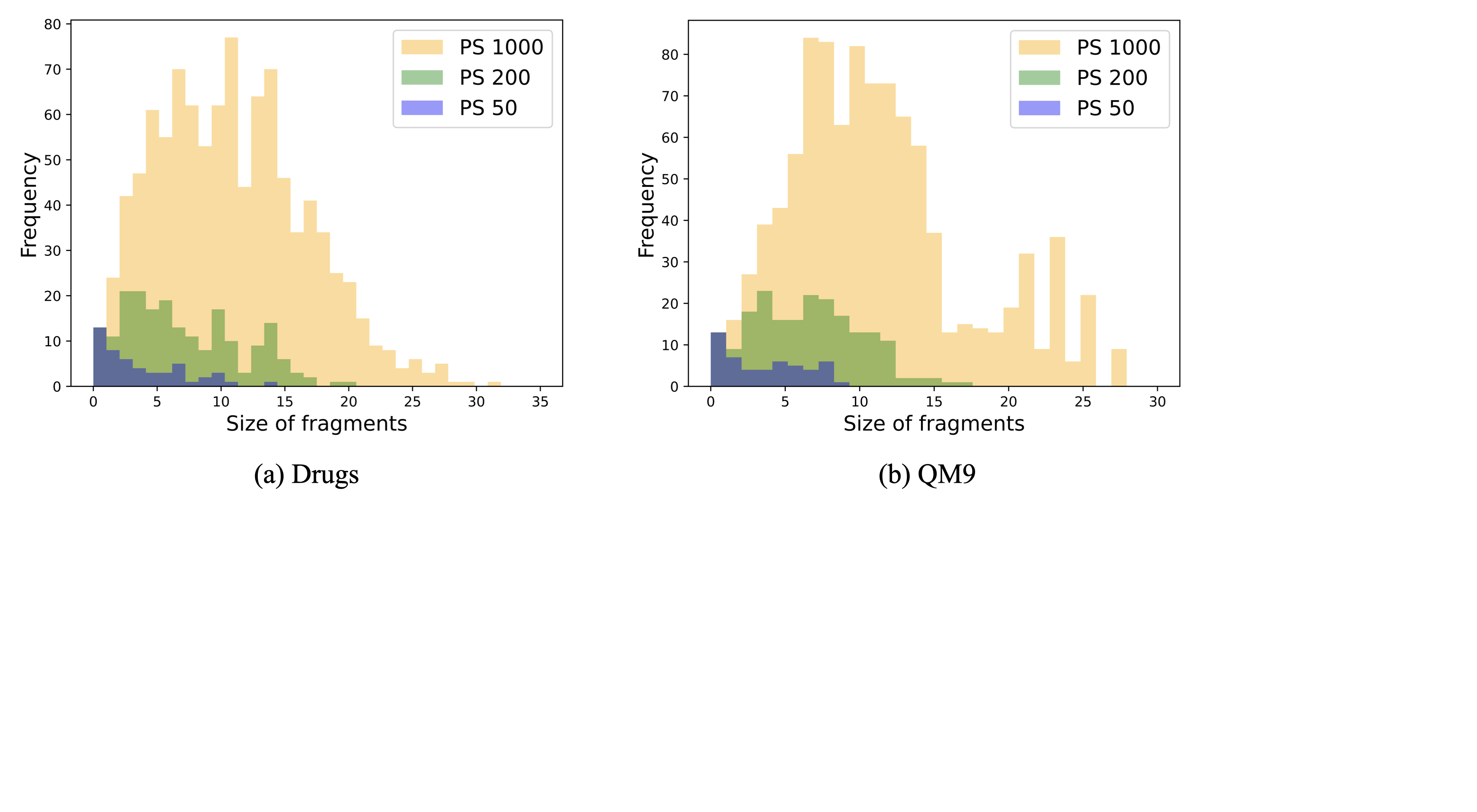}
  \caption{Frequency depends on the size of fragments in the fragment vocabulary of Drugs and QM9.}
\label{fig:vocab}
\vspace{-12pt}
\end{figure}

Additionally, the frequency depends on the size of fragments (number of constituent atoms) in Drugs and QM9 was reported in Fig. \ref{fig:vocab}. For each $\vert \mathcal{S} \vert$, the frequency distribution across fragment sizes is smooth and not biased toward certain sizes.

In the training and validation sets of the Drugs and QM9 datasets, there are 5 different ground truth conformers for each molecule. Thus, we generated 5 different conformers from RDKit to compute the fragment coordinates $\hat{\mathbf{x}}^\text{f}$ for each molecule in train and validation sets. Following \cite{jing2022torsional}, we computed the optimal matching between 5 RDKit generated conformers and 5 different ground truth conformers for a single molecule. After computing the cost matrix whose the $(i,j)$-th element means RMSD between the $i$-th RDKit generated conformers and the $j$-th ground truth conformer, we assigned optimal RDKit conformer to each ground truth conformer using linear sum assignment problem \cite{crouse2016implementing}. After finding the optimal matching, we aligned each ground truth conformer to its assigned RDKit conformer using the Kabsch algorithm \cite{kabsch1976solution}. The aligned ground truth conformers were then used in the blurring schedule (Eq. (\ref{eq:blur_sch})) and loss function (Eq. (\ref{eq:loss_gt})) of the training process.

\subsection{Training and time} \label{apdx:subsec_train}
We used a single NVIDIA A100 GPU for every training and generation tasks. For training, we used a learning rate $10^{-4}$ with the AdamW optimizer \cite{loshchilov2018decoupled}. The training time for both Drugs and QM9 was required around $3.8$ days. For sampling, Drugs required 145 minutes for 14,324 conformers in 200 molecules, and QM9 required 71 minutes for 22,408 conformers in 200 molecules. We reported hyperparameters of EBD training including the maximum time step $T$, number of layers ($\#$ $l$) and number of features ($\#$ $d$) in the deblurring networks, number of multi-hops ($\#$ of hops) and cutoff value for the expansion of atom interactions, batch size, and number of iteration in Table \ref{tab:param}.

\begin{table}[h]
\vspace{-12pt}
  \caption{Hyperparameters of EBD.}
  \label{tab:param}
  \centering
  \begin{tabular}{lccccccc}
    \toprule
    Dataset     & $T$ & $\#$ $l$ & $\#$ $d$ & $\#$ of hops &  cutoff & batch size & training iter.\\
    \midrule
    Drugs       & 50  & 6 & 128 & 3 & 10 $\text{\r{A}}$ & 32 & 650k    \\
    QM9         & 50  & 6 & 128 & 3 & 10 $\text{\r{A}}$ & 64 & 650k    \\
    \bottomrule
  \end{tabular}
\end{table}
\subsection{Performance of compared methods}
For the results of compared methods in geometric evaluation of Drugs (Table \ref{tab:rmsd_drug}) and QM9 (Table \ref{tab:rmsd_qm9}), COV-R and MAT-R scores of CVGAE \cite{mansimov2019molecular}, GraphDG \cite{simm2020generative}, CGCF \cite{xu2021learning}, and ConfGF \cite{shi2021learning} were borrowed from \cite{shi2021learning}. The performance of GeoMol and ConfVAE were borrowed from \cite{zhu2022direct} and \cite{xu2022geodiff}, respectively. In a case of RDKit \cite{landrum2013rdkit}, we reported the performance from the generated conformers from RDKit that we utilized to compute the approximate fragment coordinates $\hat{\mathbf{x}}^\text{f} \sim p_{\text{RDKit }}(\mathbf{x}^\text{f})$. For GeoDiff \cite{xu2022geodiff}, we downloaded their implementation code from \url{https://github.com/MinkaiXu/GeoDiff/tree/main} and trained GeoDiff model for our experiments. We reported the performance of GeoDiff after sampling conformers using Langevin dynamics \cite{song2019generative}, as they did in their implementation.

For the method in the ablation study, DecompDiff \cite{guan2023decompdiff} is a denoising diffusion model conditioned on coarse-grained structures, where the number of prior distributions corresponds to the number of fragments, and the mean of each prior is the respective fragment coordinates. By comparing the proposed method with DecompDiff in a controlled manner, we aimed to isolate the effect of the proposed blurring scheduler and random noise injection on learning in the coarse-to-fine molecular conformer generation task. Our use of DecompDiff was not to demonstrate its suitability for the molecular conformer generation task but rather to show that the stochastic trajectory from random noise corruption is more challenging for the coarse-to-fine generation task than the proposed blurring schedule, even when the prior distributions are conditioned on the coarse-grained structures.

\subsection{Pseudo-code}
In this subsection, we provide the Pytorch-style \cite{paszke2017automatic} pseudo-codes. The RDKit conformer generator to obtain the approximate fragment structure, linear interpolation blurring schedule, training process, and sampling process were given in Pseudo-codes \ref{code:fragmentation}, \ref{code:blurring}, \ref{code:training}, and \ref{code:sampling}, respectively. 
%\clearpage
% \begin{code}
% \captionof{Initial atom coordinate generation from RDKit.}
% \label{code:fragmentation}
% \begin{minted}[bgcolor=LightGray,]{python}
% import torch
% import copy
% from rdkit.Chem import AllChem

% def get_multiple_rdkit_coords(molecule, num_conf):
%     mol = copy.deepcopy(molecule)
%     mol.RemoveAllConformers()
%     ps = AllChem.ETDG()
%     ps.maxIterations = 5000
%     ps.randomSeed = 2023
%     ps.useBasicKnowledge = False
%     ps.useExpTorsionAnglePrefs = False
%     ps.useRandomCoords = False
%     ids = AllChem.EmbedMultipleConfs(mol, num_conf, ps)
%     if -1 in ids or mol.GetNumConformers() != num_conf:
%         print("Use DG random coords.")
%         ps.useRandomCoords = True
%         ids = AllChem.EmbedMultipleConfs(mol, num_conf, ps)
%     confs = []
%     for cid in range(num_conf):
%         confs.append(torch.tensor(mol.GetConformer(cid).GetPositions())
        
%     return confs
    
% \end{minted}
% \end{code}

\renewcommand{\lstlistingname}{Pseudo-code}
\begin{lstlisting}[language=Python, caption=Initial atom coordinate generation from RDKit., label=code:fragmentation]
import torch
import copy
from rdkit.Chem import AllChem

def get_multiple_rdkit_coords(molecule, num_conf):
    mol = copy.deepcopy(molecule)
    mol.RemoveAllConformers()
    ps = AllChem.ETDG()
    ps.maxIterations = 5000
    ps.randomSeed = 2023
    ps.useBasicKnowledge = False
    ps.useExpTorsionAnglePrefs = False
    ps.useRandomCoords = False
    ids = AllChem.EmbedMultipleConfs(mol, num_conf, ps)
    if -1 in ids or mol.GetNumConformers() != num_conf:
        print("Use DG random coords.")
        ps.useRandomCoords = True
        ids = AllChem.EmbedMultipleConfs(mol, num_conf, ps)
    confs = []
    for cid in range(num_conf):
        confs.append(torch.tensor(mol.GetConformer(cid).GetPositions())
        
    return confs
\end{lstlisting}

% \begin{listing}
% \begin{minted}[bgcolor=LightGray,]{python}

% import torch
    
% def blurring(t, x_a_gt, x_f_rdkit, mapping_matrix):
%     # prior distribution
%     x_f_rdkit_extend = mapping_matrix @ x_f_rdkit 

%     # move positions to zero center-of-mass subspace
%     x_a_gt = remove_mean(x_a_gt)
%     x_f_rdkit_extend = remove_mean(x_f_rdkit_extend)

%     # linear interpolation
%     blurred_pos = torch.lerp(x_a_gt, x_f_ref_ext_split, t) 

%     return blurred_pos
% \end{minted}
% \caption{Blurring schedule in Eq. \ref{eq:blur_sch}.}
% \end{listing}

\renewcommand{\lstlistingname}{Pseudo-code}
\begin{lstlisting}[language=Python, caption=Blurring schedule in Eq. \ref{eq:blur_sch}., label=code:blurring]
import torch
    
def blurring(t, x_a_gt, x_f_rdkit, mapping_matrix):
    # prior distribution
    x_f_rdkit_extend = mapping_matrix @ x_f_rdkit 

    # move positions to zero center-of-mass subspace
    x_a_gt = remove_mean(x_a_gt)
    x_f_rdkit_extend = remove_mean(x_f_rdkit_extend)

    # linear interpolation
    blurred_pos = torch.lerp(x_a_gt, x_f_ref_ext_split, t) 

    return blurred_pos
\end{lstlisting}
% \begin{code}
% \captionof{listing}{Training process in Algorithm \ref{alg:opt}.}
% \label{code:training}
% \begin{minted}[bgcolor=LightGray,]{python}
% import torch
    
% def loss(x_a_gt, x_f_rdkit, mapping_matrix, sigma, T):
%     # sample time
%     t = torch.randint(1, T, (1,)) / T

%     # blurred atom position from blurring schedule
%     blurred_pos = blurring(t, x_a_gt, x_f_rdkit, mapping_matrix)

%     # add noise
%     noise = torch.randn_like(blurred_pos)
%     noise = remove_mean(noise)
%     blurred_pos = blurred_pos + noise * sigma
    
%     # estimate ground truth state from blurred atom position
%     x_a_gt_estimated = deblur_network(blurred_pos, mapping_matrix, t)

%     # translate to the zero center-of-mass subspace
%     x_a_gt = remove_mean(x_a_gt)
%     x_a_gt_est = remove_mean(x_a_gt_estimated)

%     # optimal rotation matrix from Kabsch algorithm
%     rot_matrix = Kabsch_alignment(x_a_gt_est, x_a_gt)

%     # mean squared error
%     loss = mean((x_a_gt - rot_matrix @ x_a_gt_est) ** 2)
    
%     return loss
    
% \end{minted}
% \end{code}

\renewcommand{\lstlistingname}{Pseudo-code}
\begin{lstlisting}[language=Python, caption=Training process in Algorithm \ref{alg:opt}., label=code:training]
import torch
    
def loss(x_a_gt, x_f_rdkit, mapping_matrix, sigma, T):
    # sample time
    t = torch.randint(1, T, (1,)) / T

    # blurred atom position from blurring schedule
    blurred_pos = blurring(t, x_a_gt, x_f_rdkit, mapping_matrix)

    # add noise
    noise = torch.randn_like(blurred_pos)
    noise = remove_mean(noise)
    blurred_pos = blurred_pos + noise * sigma
    
    # estimate ground truth state from blurred atom position
    x_a_gt_estimated = deblur_network(blurred_pos, mapping_matrix, t)

    # translate to the zero center-of-mass subspace
    x_a_gt = remove_mean(x_a_gt)
    x_a_gt_est = remove_mean(x_a_gt_estimated)

    # optimal rotation matrix from Kabsch algorithm
    rot_matrix = Kabsch_alignment(x_a_gt_est, x_a_gt)

    # mean squared error
    loss = mean((x_a_gt - rot_matrix @ x_a_gt_est) ** 2)
    
    return loss
\end{lstlisting}
% \begin{code}
% \captionof{listing}{Sampling process in Algorithm \ref{alg:sample}.}
% \label{code:sampling}
% \begin{minted}[bgcolor=LightGray,]{python}
% import torch
% import copy 

% def sample(x_f_rdkit, mapping_matrix, delta, T):
%     # initial atom position located at fragment position
%     x_a_init = mapping_matrix @ x_f_rdkit
%     x_a_init = remove_mean(x_a_init)
%     x_a = copy.deepcopy(x_a_init)

%     for i in range(T-1, 0, -1):
%         t = i/T

%         # add noise
%         noise = torch.randn_like(x_a)
%         noise = remove_mean(noise)
%         x_a = x_a + noise * delta

%         # estimate ground truth state from blurred atom position
%         x_a_gt_est = deblur_network(x_a, mapping_matrix, t)

%         # translate to the zero center-of-mass subspace
%         x_a_gt_est = remove_mean(x_a_gt_est)

%         # optimal rotation matrix from Kabsch algorithm
%         rot_matrix = Kabsch_alignment(x_a_gt_est, x_a_init)

%         # next step from estimated ground truth and initial positions
%         x_a_gt_est = rot_matrix @ x_a_gt_est
%         x_a = blurring((i-1)/T, x_a_gt_est, x_a_init, mapping_matrx)
        
%     return x_a

% \end{minted}
% \end{code}

\renewcommand{\lstlistingname}{Pseudo-code}
\begin{lstlisting}[language=Python, caption=Sampling process in Algorithm \ref{alg:sample}., label=code:sampling]
import torch
import copy 

def sample(x_f_rdkit, mapping_matrix, delta, T):
    # initial atom position located at fragment position
    x_a_init = mapping_matrix @ x_f_rdkit
    x_a_init = remove_mean(x_a_init)
    x_a = copy.deepcopy(x_a_init)

    for i in range(T-1, 0, -1):
        t = i/T

        # add noise
        noise = torch.randn_like(x_a)
        noise = remove_mean(noise)
        x_a = x_a + noise * delta

        # estimate ground truth state from blurred atom position
        x_a_gt_est = deblur_network(x_a, mapping_matrix, t)

        # translate to the zero center-of-mass subspace
        x_a_gt_est = remove_mean(x_a_gt_est)

        # optimal rotation matrix from Kabsch algorithm
        rot_matrix = Kabsch_alignment(x_a_gt_est, x_a_init)

        # next step from estimated ground truth and initial positions
        x_a_gt_est = rot_matrix @ x_a_gt_est
        x_a = blurring((i-1)/T, x_a_gt_est, x_a_init, mapping_matrx)
        
    return x_a
\end{lstlisting}

\section{Further results on geometric evaluation} \label{apdx:further_eval}
\textbf{GEOM-QM9.} We compared our EBD to the baseline RDKit and machine learning models on small molecules GEOM-QM9, and the results are reported in Table \ref{tab:rmsd_qm9}. Compared to the most of machine learning models, EBD achieved superior performances especially on the precision score. We observed that RDKit, the distance geometry-based conformer generator, outperformed in coverage metrics for small molecules. However,  as the size of molecules increases and the tasks become more challenging, RDKit suffers a significant performance drop, as shown in Table \ref{tab:rmsd_drug}.
\begin{table}[ht]
\vspace{-12pt}
  \caption{Geometric evaluation on GEOM-QM9 benchmark ($\delta = 0.5\text{\r{A}}$).}
  \label{tab:rmsd_qm9}
  \centering
  \resizebox{0.95\textwidth}{!}{
    \begin{tabular}{l|cccc|cccc}
    \toprule[1.0pt]
     & \multicolumn{2}{c}{\shortstack[c]{COV-R ($\%$) $\uparrow$}}  & \multicolumn{2}{c|}{\shortstack[c]{MAT-R($\text{\r{A}}$) $\downarrow$}}  & \multicolumn{2}{c}{\shortstack[c]{COV-P ($\%$) $\uparrow$}}  & \multicolumn{2}{c}{\shortstack[c]{MAT-P ($\text{\r{A}}$) $\downarrow$}} \\
    % \cline{2-9}
    Models & Mean & Med & Mean & Med & Mean & Med & Mean & Med \\
    % \hline \hline
    \midrule[0.8pt]
    RDKit & 88.34 & \textbf{95.08} & 0.3544 & 0.2974 & \textbf{83.42} & \textbf{88.17} & 0.3747 & 0.3692 \\ 
    CVGAE & 0.09 & 0.00 & 1.6713 & 1.6088 & - & - & - & - \\ 
    GraphDG & 73.33 & 84.21 & 0.4245 & 0.3973 & 43.90 & 35.33 & 0.5809 & 0.5823 \\ 
    CGCF & 78.05 & 82.48 & 0.4219 & 0.3900 & 36.49 & 33.57 & 0.6615 & 0.6427 \\
    ConfVAE & 77.84 & 88.20 & 0.4154 & 0.3739 & 38.02 & 34.67 & 0.6215 & 0.6091 \\
    GeoMol & 71.26 & 72.00 & 0.3731 & 0.3731 & - & - & - & - \\ 
    ConfGF & 88.49 & 94.31 & 0.2673 & 0.2685 & 46.43 & 43.41 & 0.5224 & 0.5124 \\
    GeoDiff {\scriptsize($T=5000$)} & 88.02 & 92.33 & \textbf{0.2199} & 0.2116 & 53.72 & 52.36 & 0.4362 & 0.4259 \\
    \midrule
    EBD {\scriptsize($T=50$)}& \textbf{89.37} & 93.21 & 0.2374 & \textbf{0.1903} & 61.31 & 60.46 & \textbf{0.3622} & \textbf{0.3517} \\ 
    \bottomrule[1.0pt]
    \end{tabular}
    }
\end{table}

\begin{figure}[ht]
  \centering
  \includegraphics[trim={0 33cm 24cm 0}, clip, width=\textwidth]{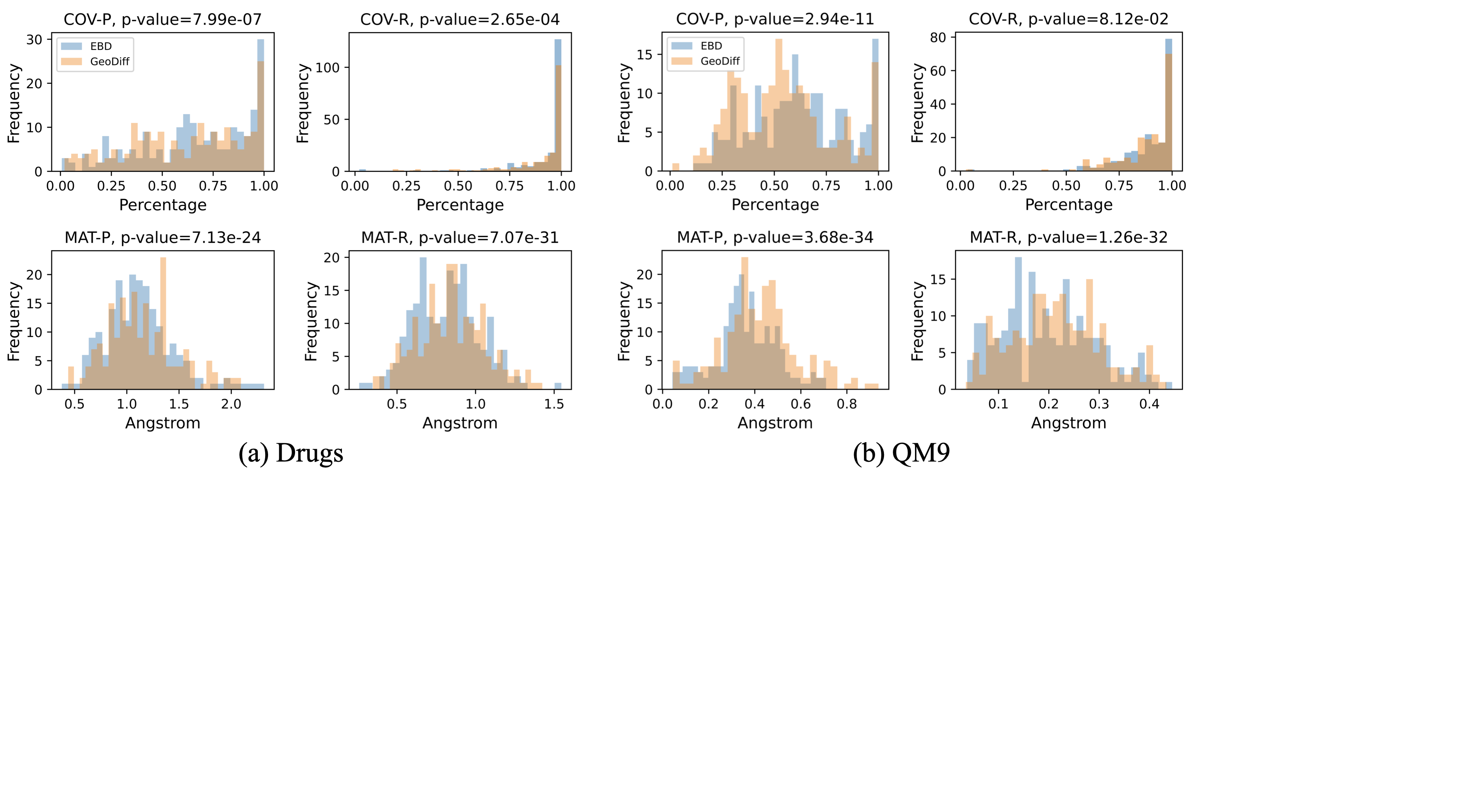}
  \caption{p-value of COV-P, COV-R, MAT-P, and MAT-R on Drugs and QM9.}
  \label{fig:stats}
    \vspace{-9pt}
\end{figure}

\textbf{Statistical significance.} We report the statistical significance of our model's improvements in geometric evaluation (COV-P, COV-R, MAT-P, and MAT-R scores). We measured p-value from one-sided Wilcoxon signed-rank test (a non-parametric version of paired t-test) over those scores of EBD and GeoDiff \cite{xu2022geodiff} on Drugs and QM9, and the results are reported in Fig. \ref{fig:stats}. Except for the COV-R score on QM9, our EBD achieved statistically significant improvement in generating more diverse and more accurate conformers for every score on either dataset, as evidenced by the p-value.
% On QM9, EBD achieved a significant improvement in accuracy, as evidenced by the p-value $< 0.05$.

\clearpage
\section{Visualizations} \label{apdx:vis}
\begin{figure}[ht]
  \centering
  \includegraphics[trim={0 8cm 8cm 0}, clip, width=\textwidth]{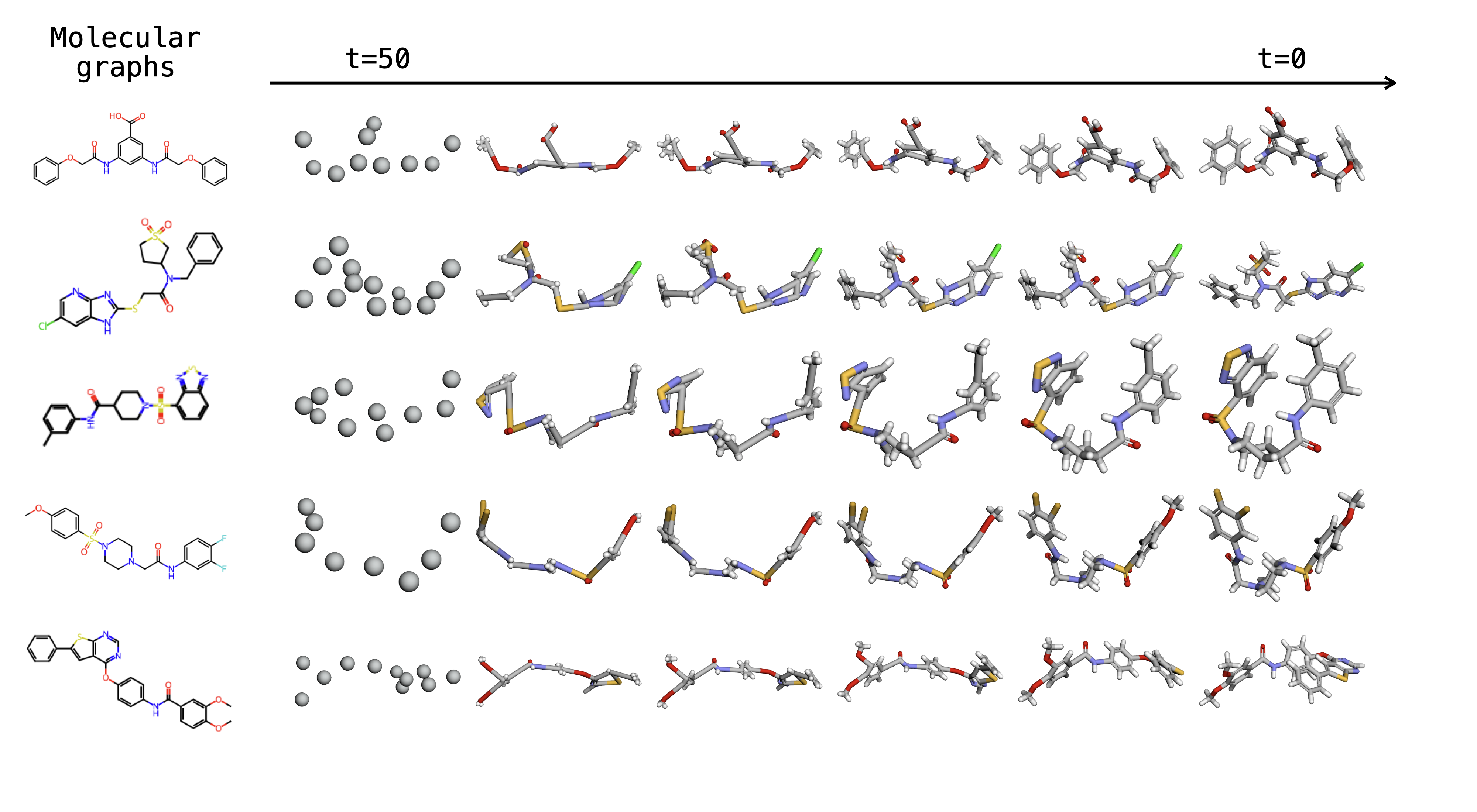}
  \caption{Sampling processes of EBD on Drugs.}
  \label{fig:traj_drug}
\vspace{-9pt}
\end{figure}

\begin{figure}[ht]
\vspace{20pt}
  \centering
  \includegraphics[trim={0 8cm 8cm 0}, clip, width=\textwidth]{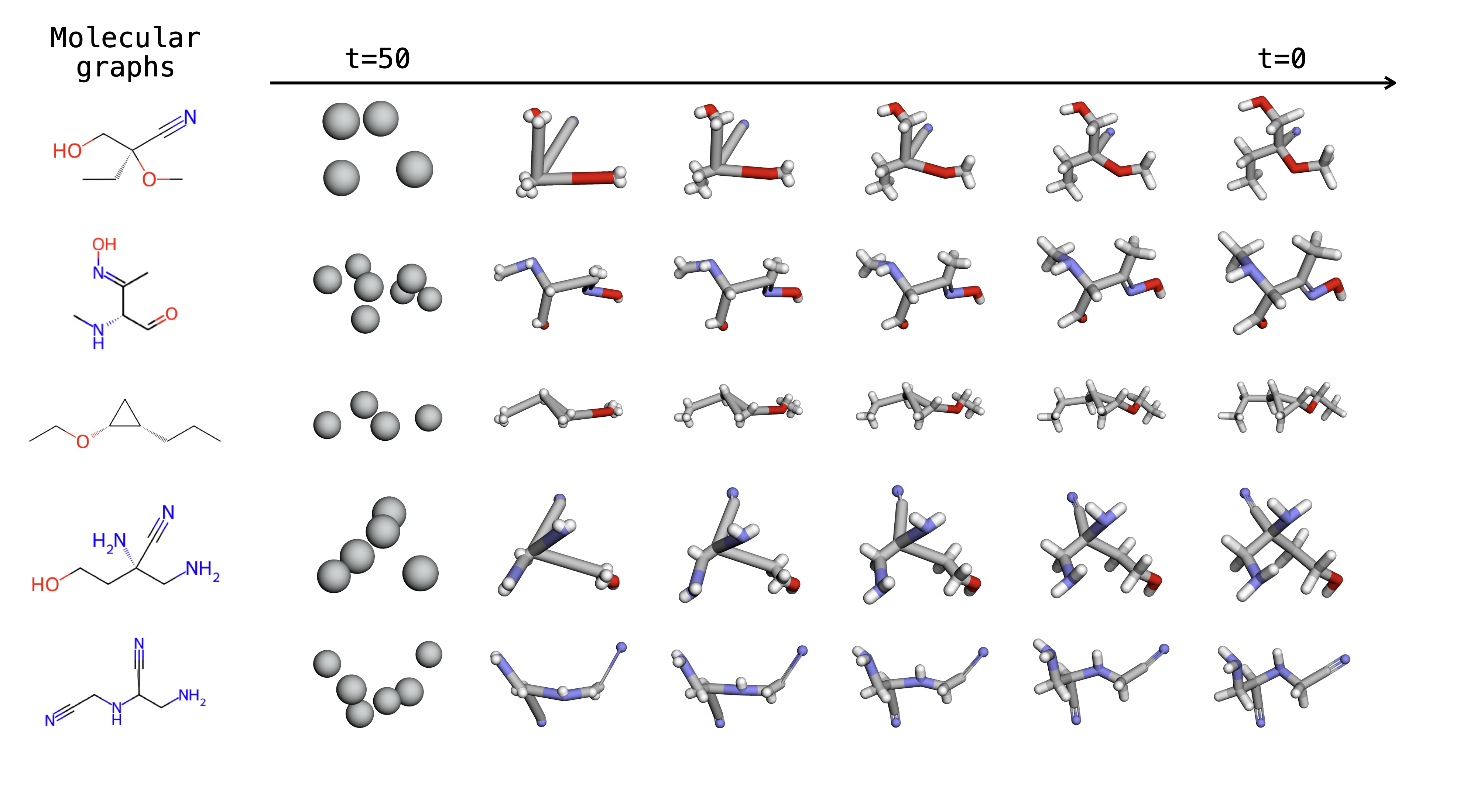}
  \caption{Sampling processes of EBD on QM9.}
  \label{fig:traj_qm9}
\vspace{-6pt}
\end{figure}

\begin{figure}[p]
  \centering
  \includegraphics[trim={0 16cm 10cm 0}, clip, width=0.95\textwidth]{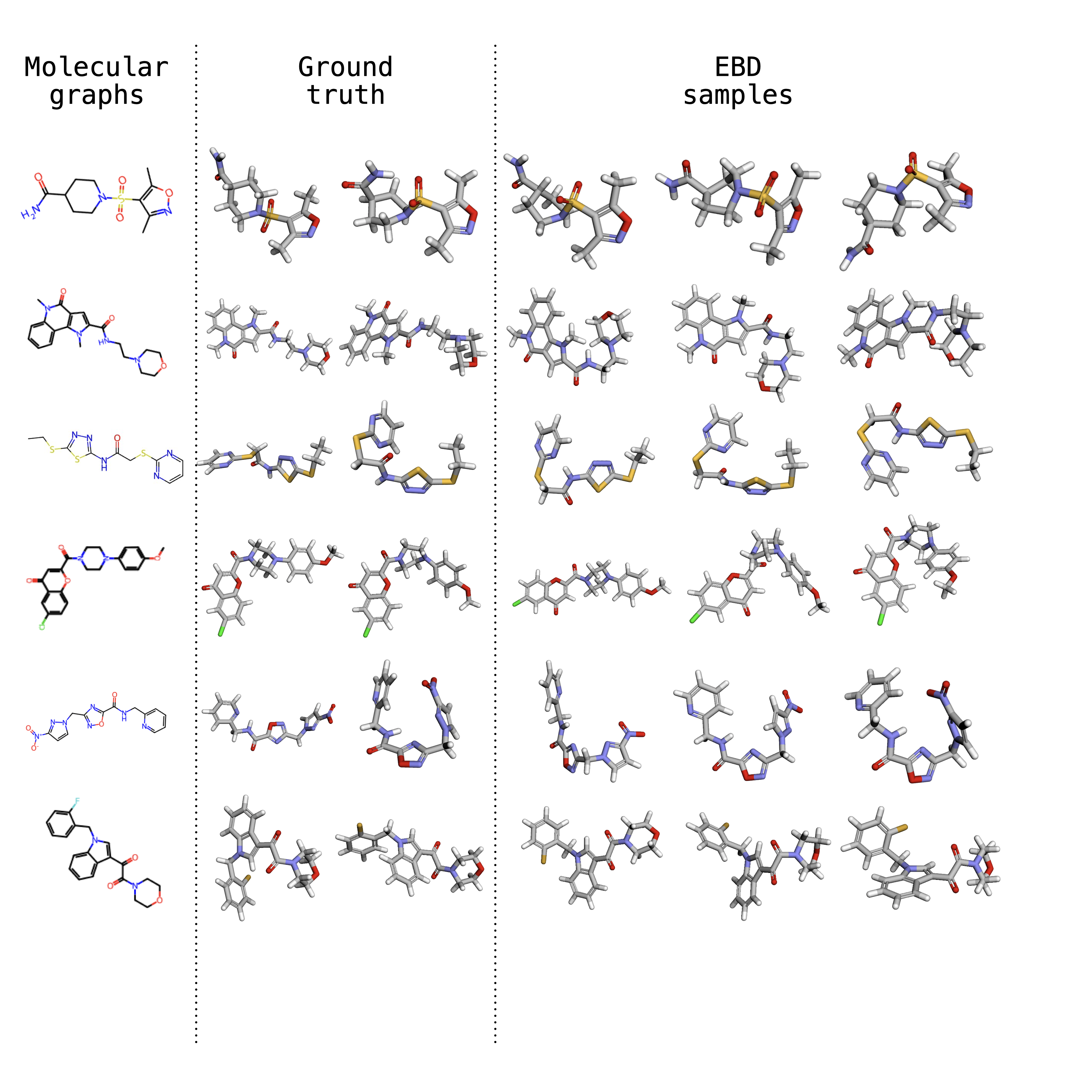}
  \caption{Visualization of molecular graphs, ground truth conformers, and samples of EBD on Drugs.}
  \label{fig:vis_drugs}
    \vspace{-9pt}
\end{figure}

\begin{figure}[p]
  \centering
  \includegraphics[trim={0 16cm 10cm 0}, clip, width=0.95\textwidth]{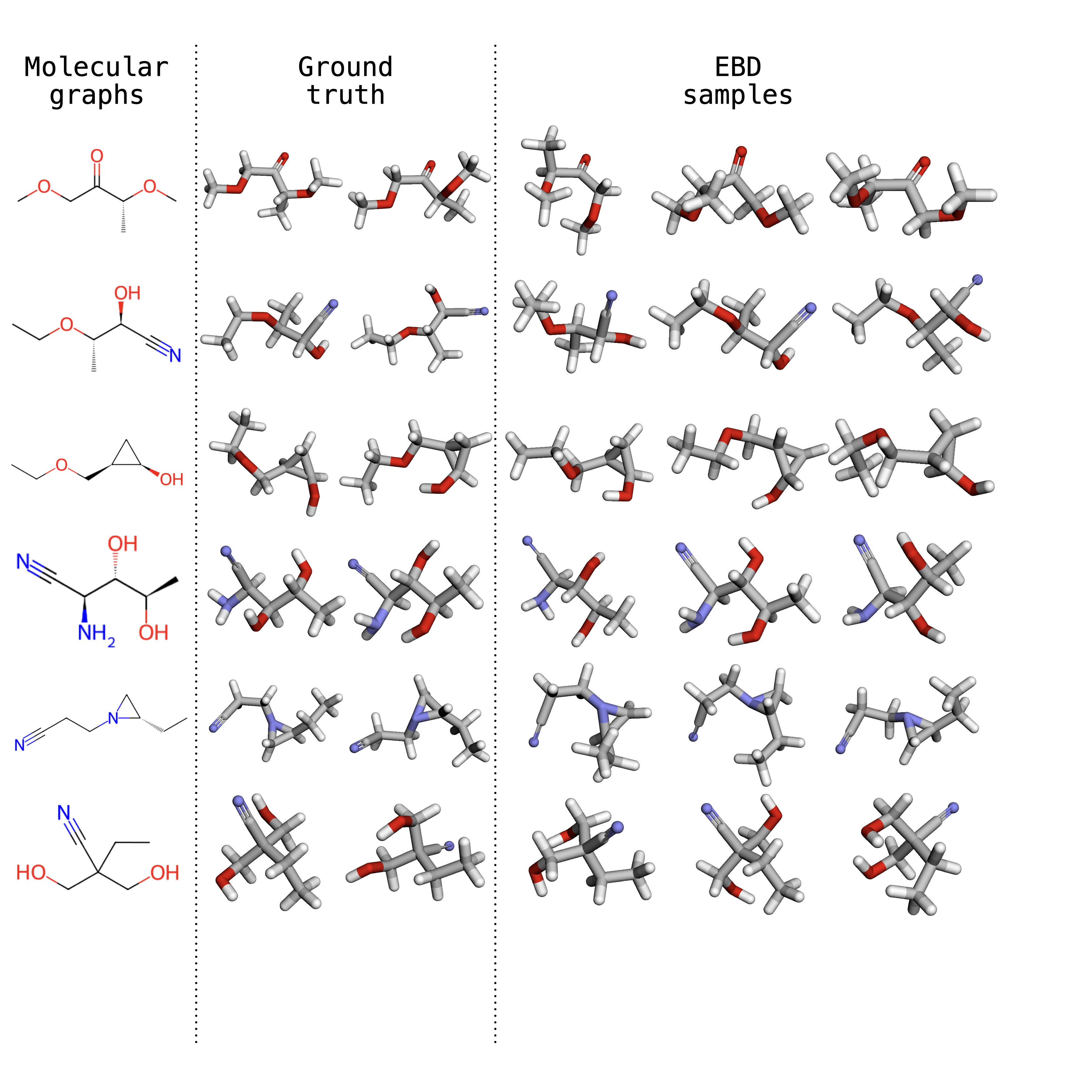}
  \caption{Visualization of molecular graphs, ground truth conformers, and samples of EBD on QM9.}
  \label{fig:vis_qm9}
\vspace{-9pt}
\end{figure}

We provide additional samples and sampling processes of EBD for the test set of Drugs in Figs. \ref{fig:traj_drug}, \ref{fig:vis_drugs} and the test set of QM9 in Figs. \ref{fig:traj_qm9}, \ref{fig:vis_qm9}.
\clearpage
\clearpage

\clearpage
\section{Limitations} \label{apdx:limitation}
Although our Equivariant Blurring Diffusion achieves significant performance on coarse-to-fine generative problems in a hierarchical molecular conformer generation scheme, there are still several limitations.

Due to the change of the estimation target from the previous state (Eq. (\ref{eq:loss_prev})) to the ground truth state (Eq. (\ref{eq:loss_gt})) during sampling (Algorithm \ref{alg:sample}), the next step $\mathbf{x}^\text{a}_{t-1}$ cannot be directly computed from the current state $\mathbf{x}^\text{a}_{t}$ and requires an additional step of the deterministic blurring function $f_{\mathbf{B}}$. This additional step in the sampling process can make the entire process slower compared to the previous state estimator.

As the size of molecule increases, the discrepancy between the ground truth $\mathbf{x}^\text{f}$ and the approximate $\hat{\mathbf{x}}^\text{f} \sim p_{\text{RDKit}}(\mathbf{x}^\text{f})$ of fragment structures becomes more severe. This increased discrepancy can make it more challenging for the model to learn the trajectory from the coarse fragment structures to the fine atomic details. To circumvent this issue, increasing the time step $T$ to more than 50 can be applied. Also, for a more accurate deblurring network than equivariant graph neural networks \cite{satorras2021n} we used, more powerful geometric graph neural networks \cite{joshi2023expressive} can be applied such as local complete frames \cite{du2022se} and higher-order tensors from spherical harmonics \cite{thomas2018tensor}.

\section{Broader impacts} \label{apdx:impacts}
We presented a deep generative model for the coarse-to-fine generation of molecular conformers. Our proposed model can be applied to problems in fragment-based drug discovery, such as scaffold hopping and linker generation, to achieve improved performance. In drug discovery applications, potential negative societal impacts may arise if the training set is contaminated and includes toxins. In such cases, the generated samples could potentially be harmful to humans. From a more general perspective, the generative models to which our model belongs can be misused to create false information that appears authentic. Therefore, users must be aware of the potential risks associated with generative models before using them.

\end{document}